\documentclass[journal]{IEEEtran}
\usepackage[T1]{fontenc}
\usepackage[utf8]{inputenc}
\usepackage{color}
\usepackage{array}
\usepackage{pifont}
\usepackage{multirow}
\usepackage{amsmath}
\usepackage{amssymb}
\usepackage{graphicx}
\PassOptionsToPackage{normalem}{ulem}
\usepackage{ulem}

\makeatletter

\providecommand{\tabularnewline}{\\}

\let\oldforeign@language\foreign@language
\DeclareRobustCommand{\foreign@language}[1]{%
  \lowercase{\oldforeign@language{#1}}}

\newcommand{\ignore}[1]{}
\usepackage{epsfig}
\usepackage{bm}
\usepackage{cite}
\usepackage{array}
\usepackage{extarrows}
\usepackage{url}
\usepackage{makecell}

\usepackage{multirow}
\usepackage{booktabs}
\usepackage{blkarray}
\usepackage{multirow}
\usepackage{array}
\usepackage{pmat}

\usepackage[font=normalsize]{caption}

\usepackage{etoolbox}

\patchcmd{\@makecaption}
{\scshape}
{}
{}
{}

\patchcmd{\@makecaption}
{\\}
{.\ }
{}
{}

\usepackage[switch]{lineno}

\makeatother

\begin{document}
\title{Hierarchical Paired Channel Fusion Network for Street Scene Change Detection}
\author{Yinjie Lei, Duo Peng, Pingping~Zhang$^{*}$, Qiuhong Ke and~Haifeng~Li
\thanks{Copyright (c) 2020 IEEE. Personal use of this material is permitted.
However, permission to use this material for any other purposes must
be obtained from the IEEE by sending an email to \textcolor{blue}{\underline{pubs-permissions@ieee.org}}.

$^{*}$The corresponding author.

Y. Lei and D. Peng are with College of Electronics and Information
Engineering, Sichuan University, Chengdu, Sichuan 610065, China. (Email: yinjie@scu.edu.cn and
duo\_peng@stu.scu.edu.cn)

P. Zhang is with School of Artificial Intelligence, Dalian University
of Technology, Dalian, Liaoning 116024, China. (Email: zhpp@dlut.edu.cn)

Q. Ke is with School of Computing and Information Systems, University
of Melbourne, Parkville, VIC 3052, Australia. (Email: qiuhong.ke@unimelb.edu.au)

H. Li is with School of Geosciences and Info-Physics, Central South University,
Changsha, Hunan 410012, China. (Email: lihaifeng@csu.edu.cn)

This work is supported in part by the National Natural Science Foundation
of China (NNSFC), No. 61403265, No. 41571397, No. 61725202, No. 61751212
and No. 61771088. This work is also supported in part by the Key Research
and Development Program of Sichuan Province (2019YFG0409).}}
\markboth{IEEE Transactions on Image Processing}{}
\maketitle
\begin{abstract}
Street Scene Change Detection (SSCD) aims to locate the changed regions
between a given street-view image pair captured at different times,
which is an important yet challenging task in the computer vision
community. The intuitive way to solve the SSCD task is to fuse the
extracted image feature pairs, and then directly measure the dissimilarity
parts for producing a change map. Therefore, the key for the SSCD
task is to design an effective feature fusion method that can improve
the accuracy of the corresponding change maps. To this end, we present
a novel Hierarchical Paired Channel Fusion Network (HPCFNet), which
utilizes the adaptive fusion of paired feature channels. Specifically,
the features of a given image pair are jointly extracted by a Siamese
Convolutional Neural Network (SCNN) and hierarchically combined by
exploring the fusion of channel pairs at multiple feature levels.
In addition, based on the observation that the distribution of scene
changes is diverse, we further propose a Multi-Part Feature Learning
(MPFL) strategy to detect diverse changes. Based on the MPFL strategy,
our framework achieves a novel approach to adapt to the scale and
location diversities of the scene change regions. Extensive experiments
on three public datasets (i.e., PCD, VL-CMU-CD and CDnet2014) demonstrate
that the proposed framework achieves superior performance which outperforms
other state-of-the-art methods with a considerable margin.
\end{abstract}


\IEEEpeerreviewmaketitle{ }

\section{Introduction}

\begin{figure}
\centering{}\centering{}\centering \resizebox{0.5\textwidth}{!}{
\begin{tabular}{@{}c@{}c@{}c}
\includegraphics[width=0.3\linewidth,height=3.6cm]{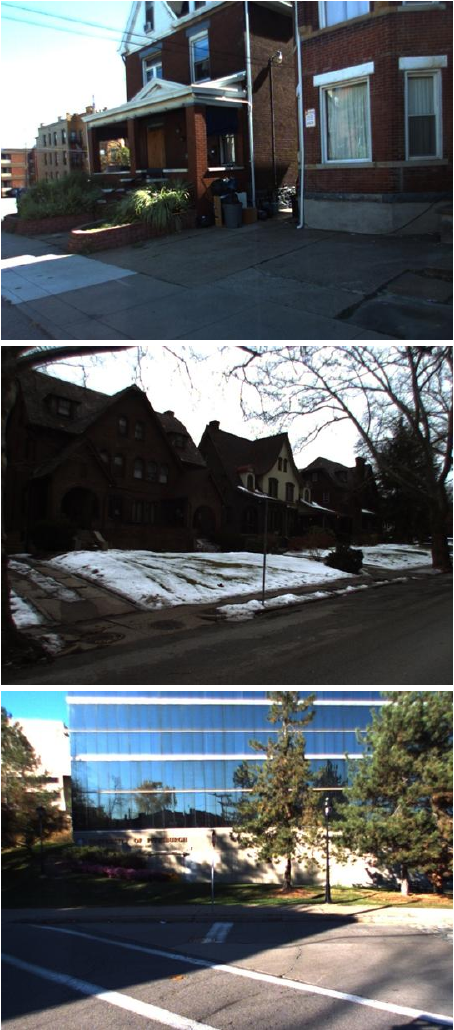}  & 
\includegraphics[width=0.3\linewidth,height=3.6cm]{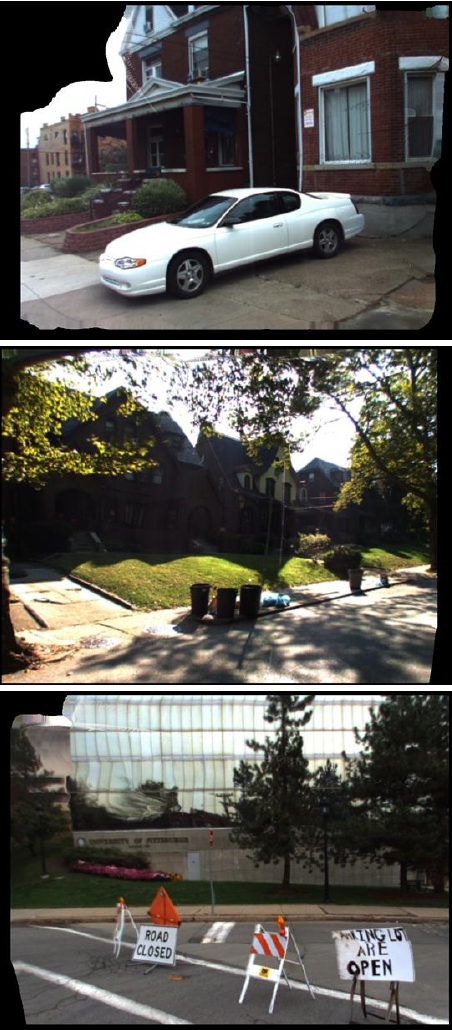}  & 
\includegraphics[width=0.3\linewidth,height=3.6cm]{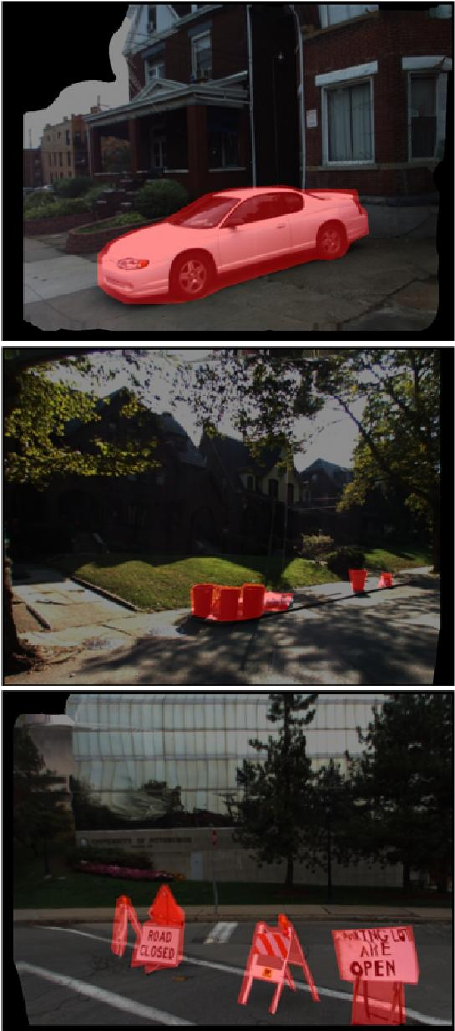} \tabularnewline
{\small{}(a) Image at $t_{0}$}  & {\small{}(b) Image at $t_{1}$}  & {\small{}(c) Change map} \tabularnewline
\end{tabular}} \caption{Examples of the SSCD task. (a) The image at time $t_{0}$; (b) The
image at time $t_{1}$; (c) The predicted change map which are highlighted
in red.\label{fig:Examples-of-Scene}}
\vspace{-4mm}

\end{figure}

\IEEEPARstart{S}{cene} Change Detection (SCD) aims to find the changed
regions between a pair of images captured at different times. This
task has been attracting growing research interest as its various
real-world applications, e.g., global resources monitoring, land-use
change detecting, disaster evaluation, visual surveillance and urban
management. In recent years, two SCD application scenarios have seen
a lot of research activities, i.e., Remote Sensing and Street Surveillance.
Many researchers \cite{wu2017kernel,wu2016scene,hussain2013change,mou2018learning,du2018unsupervised,wu2013slow}
have made efforts on the topic of Remote-Sensing SCD (RSSCD) for the
purpose of information analysis on earth's surface. Meanwhile, important
contributions \cite{sakurada2017dense,sakurada2018weakly,guo2018learning,khan2016learning,sakurada2015change}
have been devoted to Street Scene Change Detection (SSCD) aiming to
overcome the issues of urban management. In this paper, we focus on
SSCD. A sample illustration of the SSCD task is shown in Fig. \ref{fig:Examples-of-Scene}.
The fundamental method for the SSCD task is to combine the extracted
features from the input image pair, and then the SSCD is cast as a
pixel-wise classification task for generating a change map. Intuitively,
the information fusion from different sources plays an important role
in the SSCD task since the change map is jointly determined by the
input image pair. During the past decades, lots of SSCD methods have
been proposed \cite{rosin1998thresholding,rosin2003evaluation,pollard2007change,schindler2010probabilistic,taneja2011image,taneja2013city,hussain2013change}.
Most of these methods adopt handcrafted features as detection cues.
However, the handcrafted features suffer from some inherent drawbacks,
e.g., low robustness and insufficient semantics, which significantly
hinder the performance of the SSCD task. Recently, deep Convolutional
Neural Networks (CNNs) \cite{long2015fully,ronneberger2015u,szegedy2015going}
have been successfully applied to various pixel-wise image classification
tasks due to their great capabilities to extract multi-level feature
representations. Encouraged by their strengths, researchers \cite{sakurada2017dense,alcantarilla2018street,khan2016learning,sakurada2015change,sakurada2018weakly,guo2018learning}
start to leverage CNNs to automatically fuse features from different
sources for the SSCD task in order to avoid the intrinsic drawbacks
of handcrafted features.

\begin{figure*}[ht]
\begin{centering}
\begin{tabular}{@{}c@{}c@{}c@{}c@{}c@{}c@{}c}
\includegraphics[width=0.23\linewidth,height=4.3cm]{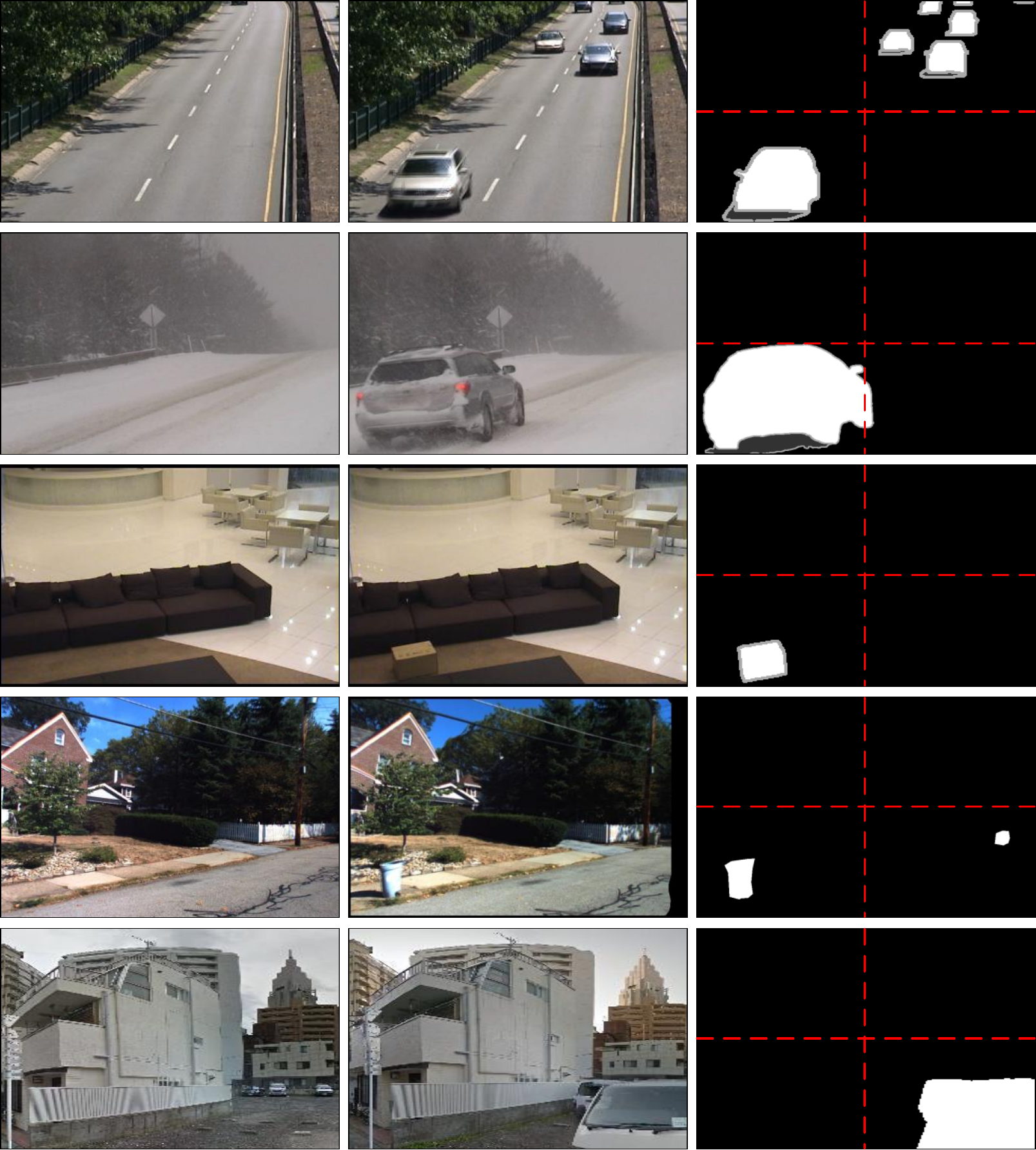}  &  & \includegraphics[width=0.23\linewidth,height=4.3cm]{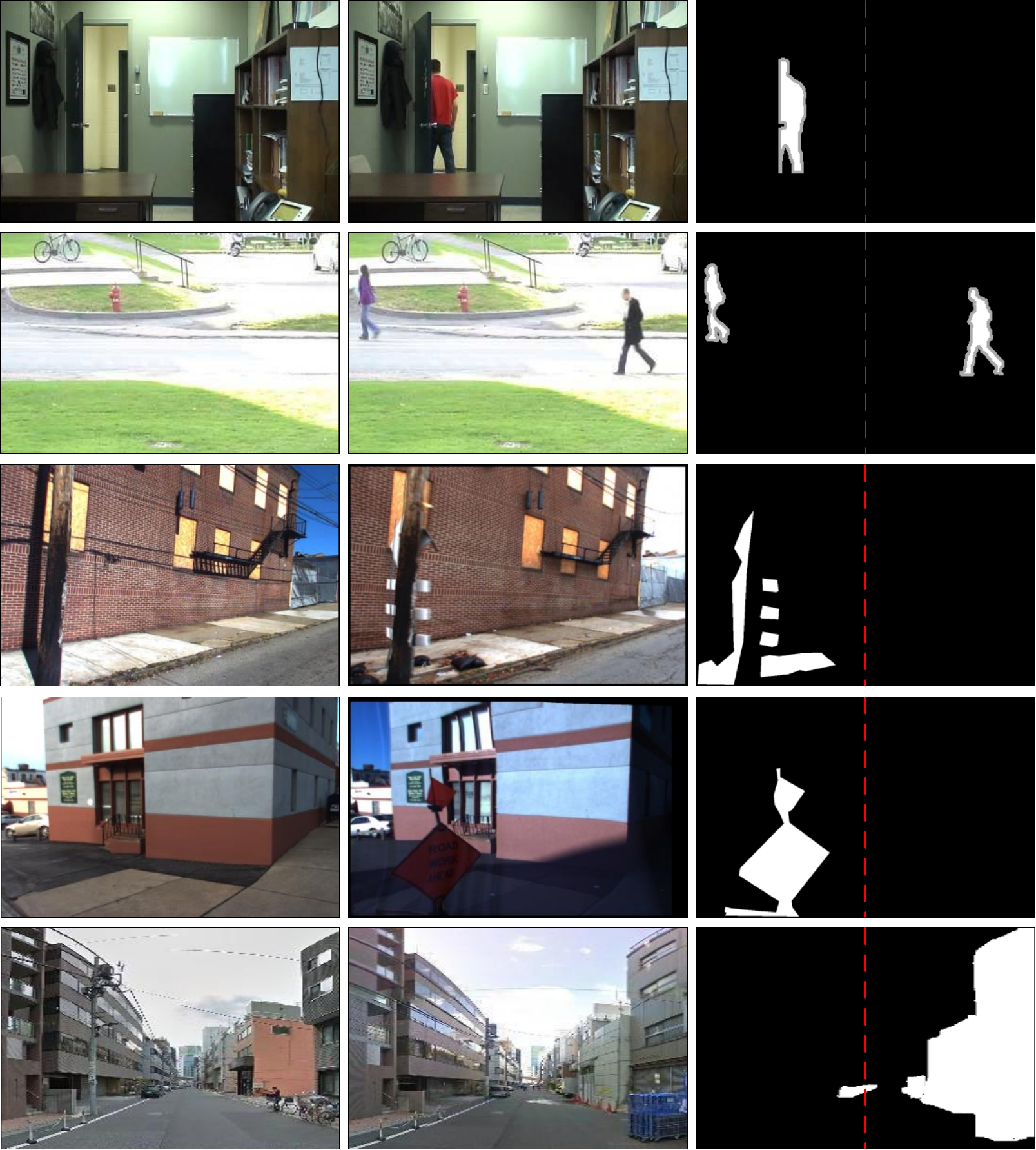}  &  & \includegraphics[width=0.23\linewidth,height=4.3cm]{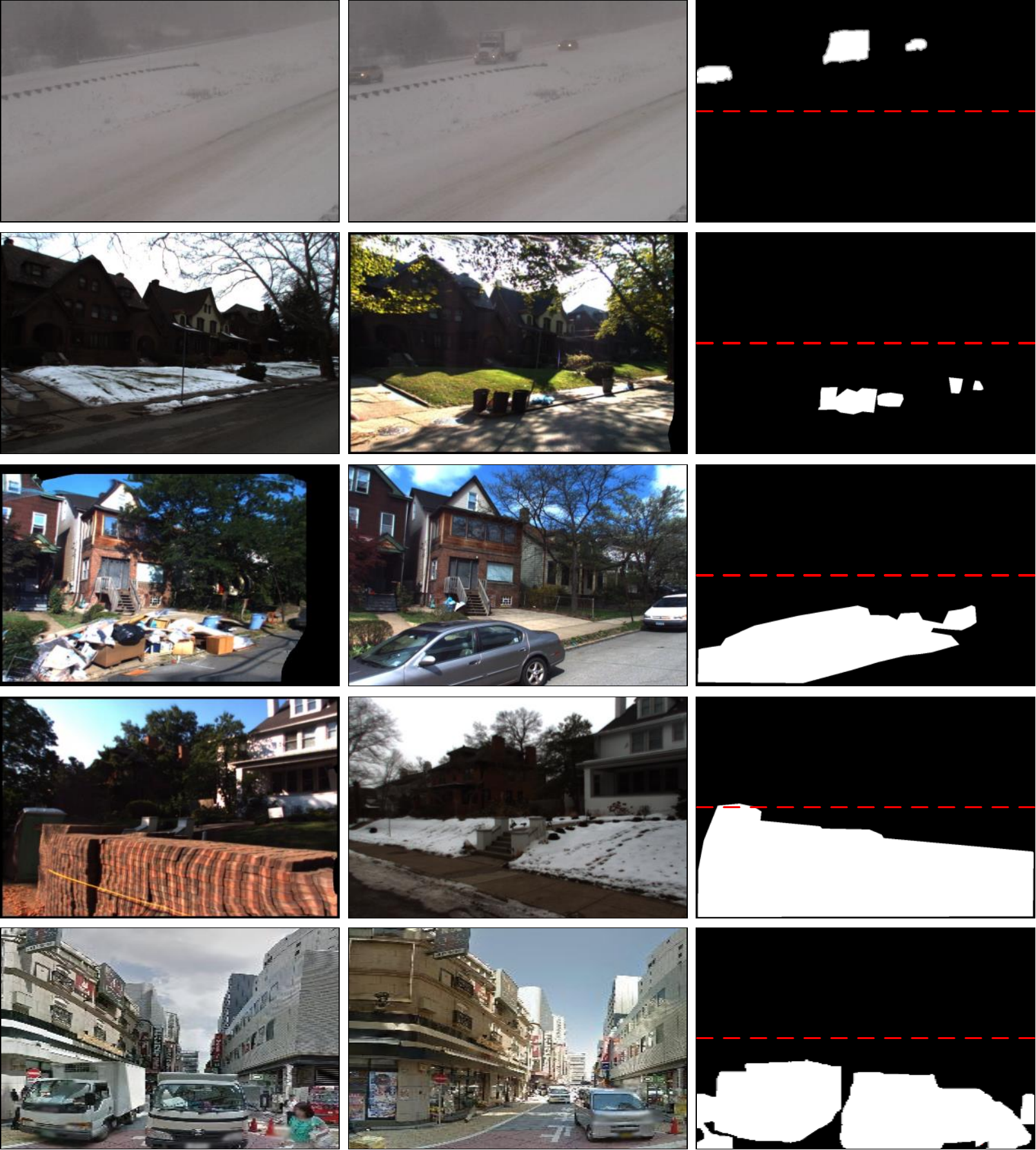}  &  & \includegraphics[width=0.23\linewidth,height=4.3cm]{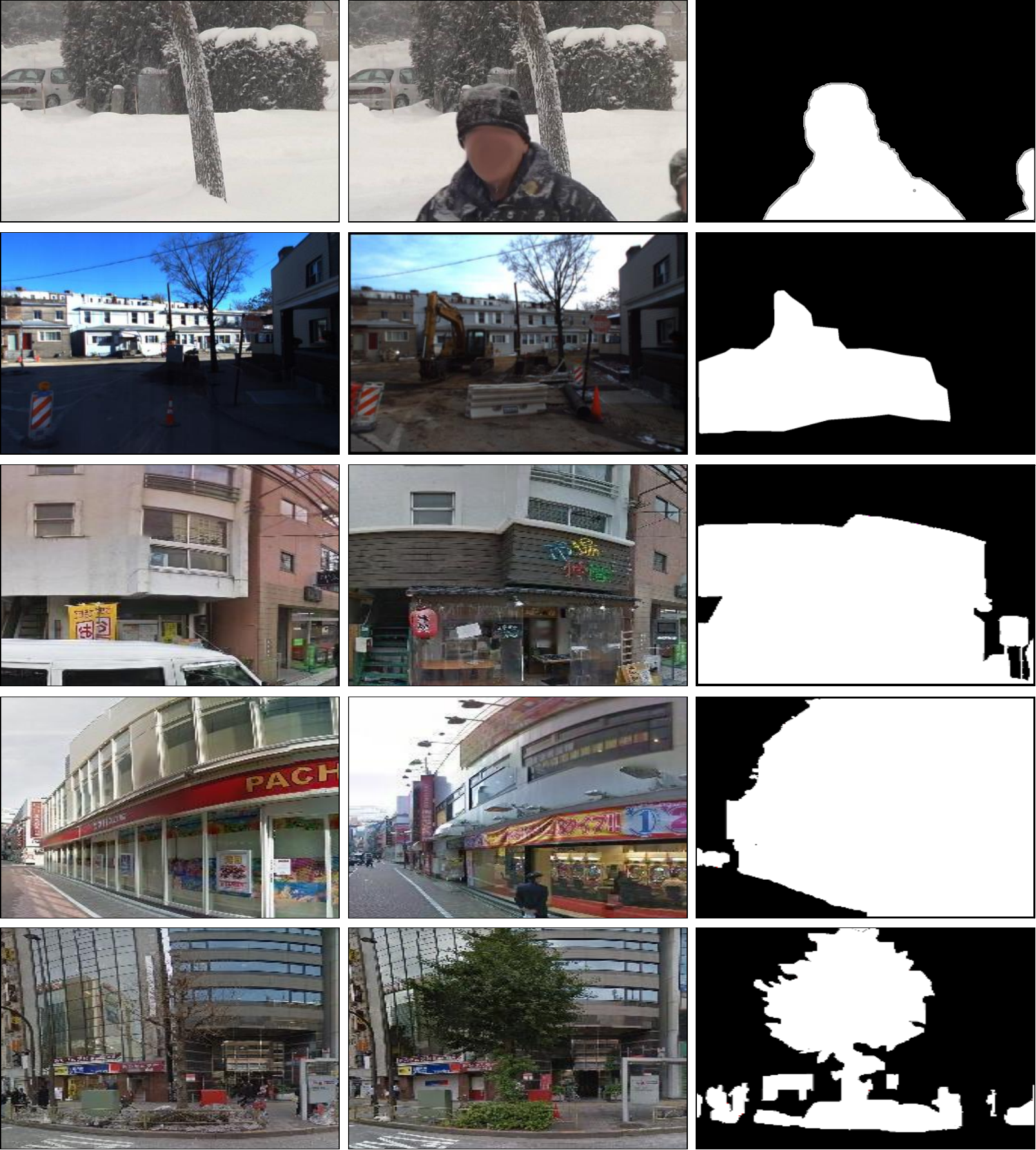}\tabularnewline
{\small{}(a) Corner}  & \,\,\,\,  & {\small{}(b) Left and right}  & \,\,\,\,  & {\small{}(c) Top and bottom}  & \,\,\,\,  & {\small{}(d) Center}\tabularnewline
\end{tabular}
\par\end{centering}
\caption{Examples from three widely-used datasets: PCD, VL-CMU-CD and CDnet2014.
We can see that the locations and scales of changed regions are unbalanced
in the dataset. It can be divided into four cases: (a) corner, (b)
left and right, (c) top and bottom and (d) center.\label{fig:Examples-from-three}}
\end{figure*}

\begin{figure}
\centering{}\centering{}\centering \resizebox{0.5\textwidth}{!}{
\begin{tabular}{@{}c}
\includegraphics[width=1\linewidth,height=2.6cm]{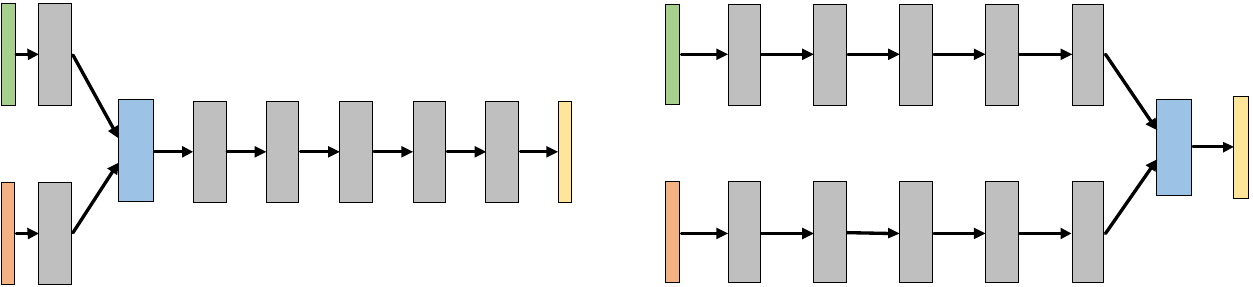}\tabularnewline
{\small{}{}(a) Single-level feature fusion: Early Fusion (left) and
Late Fusion (right).}\tabularnewline
\includegraphics[width=1\linewidth,height=2.2cm]{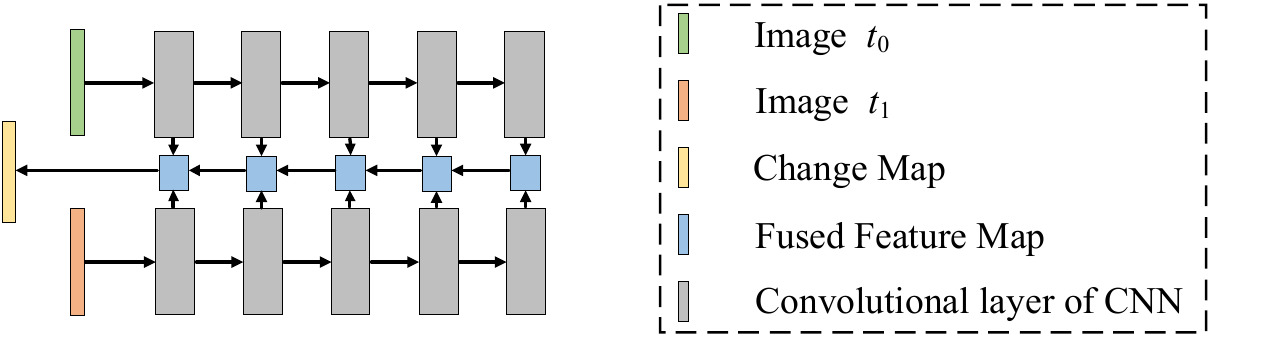}\tabularnewline
{\small{}{}(b) Multi-level feature fusion: Dense Fusion.}\tabularnewline
\end{tabular}} \caption{Existing fusion structures for the SSCD task. (a) Early Fusion and
Late Fusion are adopted in single-level fusion. (b) Dense Fusion is
adopted in multi-level fusion.\label{fig:Traditional-architectures}}
\vspace{-4mm}

\end{figure}

As shown in Fig. \ref{fig:Traditional-architectures}, the CNN based
SSCD methods can be coarsely classified into two groups: single-level
feature fusion methods, and multi-level feature fusion methods. In
the past few years, many methods \cite{sakurada2017dense,alcantarilla2018street,khan2016learning,guo2018learning,sakurada2015change}
adopt single-level feature based architecture, e.g., early-fusion
or late-fusion (Fig. \ref{fig:Traditional-architectures} (a)), which
fuse the features from different sources at a specific fusion position.

However, the single-level feature methods can only fuse partial information
for the SSCD task, which may hinder the detection performance. In
contrast, the multi-level feature methods are capable of representing
different characteristics of change scenes, which can significantly
improve the performance. Therefore, several multi-level feature based
methods have been proposed \cite{sakurada2018weakly,zhang2020rapnet},
which perform the dense feature fusion (shown in Fig. \ref{fig:Traditional-architectures}
(b)) in a deep-to-shallow manner. However, most of existing multi-level
feature fusion methods follow a simple fusion manner, i.e., concatenation
or summation operation. As many visualization works shown, the features
at shallow layers of deep networks contain low-level vision information,
such as image texture, object boundary, scene edge, etc. With the
increase of layers, the hierarchical features will become more abstractive.
In general, the deep layers have high-level information, such as image
content, semantic concept, etc. Thus, there are semantic gaps at different
levels of network layers. Such a naive fusion ignores the semantic
relationship between different feature maps.

To address the above problems, we propose a novel Hierarchical Paired
Channel Fusion Network (HPCFNet), which is a more effective framework
for the multi-level feature fusion. Specifically, for each feature
level, we introduce a Paired Channel Fusion (PCF) module which enables
the cross-image feature fusion to sufficiently capture the channel-wise
changes. The paired image features are jointly extracted by a Siamese
Convolutional Neural Network (SCNN), and then hierarchically combined
in a coarse-to-fine manner. In addition, a Reverse Spatial Attention
(RSA) mechanism is presented to highlight the changed regions while
suppressing the unchanged regions.

Besides, we observe that both the locations and scales of changed
regions are diverse in the dataset. As shown in Fig. \ref{fig:Examples-from-three},
the spatial distribution of changed regions is diverse and changed
regions contain various scales. For example, the changed regions in
Fig. \ref{fig:Examples-from-three} (a) are located in the corner
and have small scales, while those in Fig. \ref{fig:Examples-from-three}
(d) are located in the center with large scales. The imbalance can
be divide into 4 cases: (a) corner, (b) left and right, (c) top and
bottom and (d) center. Based on the above observations, we propose
a Multi-Part Feature Learning (MPFL) strategy, which adapts to the
location and scale variations. More specifically, the MPFL strategy
consists of four branches with different partition methods. Four branches
utilize adaptive convolutional layers to capture the discriminative
characteristics from different spatial parts. The proposed framework
has been evaluated on three large-scale SSCD datasets, i.e., PCD \cite{sakurada2015change},
VL-CMU-CD \cite{alcantarilla2018street} and CDnet2014 \cite{wang2014cdnet}.
Experimental results have demonstrated that the proposed framework
outperforms other state-of-the-art methods with a considerable margin.

Our contributions can be summarized as follows:
\begin{itemize}
\item We propose a novel framework, named HPCFNet for the challenging SSCD
task. The HPCFNet efficiently utilizes a dense fusion architecture
for multi-level feature fusion. Besides, an effective RSA module
is incorporated to highlight changed regions from the fused feature
maps.
\item We propose a novel MPFL strategy to detect changed regions from holistic
to local. The MPFL addresses the spatial distribution and diverse
scales of changed regions by using four different partition methods.
\item Comprehensive experiments on three public SSCD datasets demonstrate
that the proposed framework achieves superior performance and outperforms
other state-of-the-art methods with a considerable margin.
\end{itemize}
The reminder of this work is organized as follows. Section \ref{sec:Related-Works}
presents an overview of related works on change detection, feature
fusion and attention mechanism. Section \ref{sec:Methodology} describes
our proposed method in detail. The experimental setups and results
are provided in Section \ref{sec:Experimental-Setups} and Section
\ref{sec:Experimental-Results}, respectively. The conclusions are
drawn in Section \ref{sec:Conclusion}.

\section{Related Works\label{sec:Related-Works}}

\subsection{Change Detection\label{subsec:Change-detection}}

In the past few years, with the development of semantic segmentation
\cite{long2015fully,zhang2020deep,zhang2020rapnet,zhang2019hyperfusion,zhang2019cascaded,zhang2018salient,ma2020global},
many SSCD methods have been proposed. The technical comparisons of
different handcrafted feature based methods are summarized in \cite{radke2005image}.
A detailed review is beyond the scope of this work. In this section,
we mainly describe SSCD methods based on deep learning.

In recent years, deep neural networks have been successfully applied
in many research fields and have achieved the most advanced performance.
Almost all the excellent SSCD methods are based on pre-trained CNN
backbones~\cite{he2016deep,szegedy2015going,krizhevsky2012imagenet,lecun1998gradient,simonyan2014very,liu2020semi,lei2021towards},
and many of them are based on Fully Convolutional Networks (FCN) \cite{long2015fully}.
For example, Sakurada et al.~\cite{sakurada2015change} calculated
the distance between features extracted from paired CNNs, and eliminated
the geometric background context by superpixel segmentation. Guo et
al. \cite{guo2018learning} proposed to learn the distinguishing features
with the customized feature distance metrics. Zhan et al. \cite{zhan2017change}
proposed a contrastive loss function for CNNs, which increases the
distance of the feature points which are identified as changed and
reduces the distance of the feature points which are identified as
unchanged. Sakurada et al. \cite{sakurada2017dense} utilized dense
optical flow and CNNs to model the spatial correspondences between
images captured at different times. Alcantarilla et al. \cite{alcantarilla2018street}
used the dense geometry and accurate registration to warp images from
different times for the change comparison. Khan et al. \cite{khan2016learning}
proposed a deep CNN with Directed Acyclic Graph (DAG) topologies to
measure image changes. Sakurada et al. \cite{sakurada2018weakly}
introduced hierarchically dense connections to capture the multi-scale
feature information.

Although previous methods achieve remarkable performance for the SSCD
task, they still have several unsolved problems. Most of the recent
deep learning methods focus on adding auxiliary information, designing
loss functions or using additional data. However, few of them have
focused on further improvement of feature fusion, especially fusion
based on multi-level features. To solve this problem, we introduce
a novel feature fusion network to hierarchically exploit paired channel
differences of multi-level features.

\subsection{Feature Fusion\label{subsec:Feature-Fusion}}

Encouraged by the remarkable strengths of deep CNNs, a few methods
for the SSCD task leverage CNNs to fuse information from different
image sources. In \cite{alcantarilla2018street,khan2016learning,sakurada2017dense},
the information from two sources was combined at a shallow position.
In contrast, late fusion based methods \cite{guo2018learning,sakurada2015change}
fuse the two source features at a deep position. Essentially, the
early fusion and late fusion can be merged as a single-level feature
fusion strategy. Though the single-level feature fusion strategy achieves
encouraging performance, it does not utilize all the available information.
Due to the powerful ability of presentations, the features at other
levels are also important for the SSCD task.

To take advantage of multi-level features, the method in \cite{sakurada2018weakly}
has been introduced by using dense fusion positions inside the networks\cite{zhang2019hyperfusion,zhang2018salient,zhang2019salient,zhang2019cascaded,zhang2020deep}.
However, this method still follows the traditional fusion manner such
as concatenation and element-wise addition to combine multiple features.
In order to address this problem, we propose a novel Paired Channel
Fusion (PCF) module to sufficiently fuse the channel pairs at each
feature level. In the PCF module, we utilize atrous convolutions with
various dilation rates to generate diverse fused features.

\subsection{Attention Mechanism\label{subsec:Attention-based-on}}

Attention mechanisms have been developed over many decades in neuroscience
community \cite{itti2001computational,itti1998model,olshausen1993neurobiological,zhang2019deep}
and have played a vital role in current deep neural networks \cite{larochelle2010learning,mnih2014recurrent}.
Previous works in \cite{mnih2014recurrent,jaderberg2015spatial,zhang2019deep,fu2019dual,chen2018reverse}
have shown the importance of attention mechanism, which uses high-level
information to weigh features at the middle of networks. Attention
usually cooperates with gating functions (e.g., softmax and sigmoid
function), and has achieved excellent performance for sequence and
image localization \cite{cao2015look,jaderberg2015spatial}. Recently,
Hu et al. \cite{hu2018squeeze} proposed a self-attention unit called
the Squeeze-and-Excitation (SE) block. The goal of the SE block is
to ensure that the network can highlight related features and suppress
the less useful features. Woo et al. \cite{woo2018cbam} proposed
the Convolutional Block Attention Module (CBAM) to intensify the meaningful
information on both channel and spatial axis. Channel and spatial
attention modules are jointly utilized to learn task-related features
of multiple branches. Inspired by previous works, we propose a Reverse
Spatial Attention (RSA) module to generate an attention mask by pooling
operations (average pooling and max pooling) and feature reversing.
The RSA highlights the changed regions based on the fused feature
maps of two extracted feature maps.
\begin{figure*}[!th]
\begin{centering}
\includegraphics[width=0.8\linewidth,height=7cm]{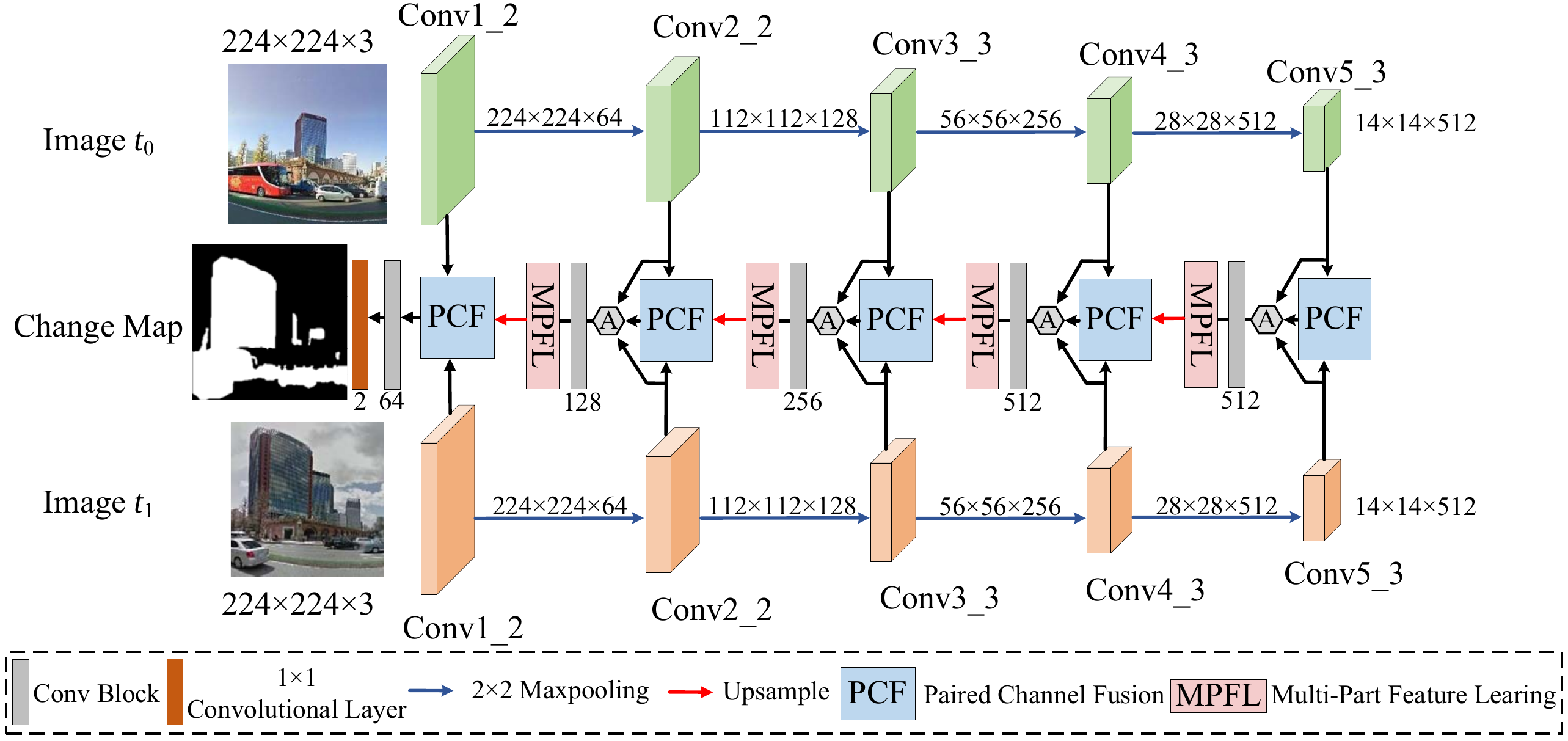}
\par\end{centering}
\caption{The overall architecture of the HPCFNet. ``A'' denotes the Reverse
Spatial Attention (RSA) module. Paired Channel Fusion (PCF) modules
are used for channel-wise feature fusion. The Multi-Part Feature Learning
(MPFL) integrates the features from holistic to local scales, and
adapt to various change distributions.\label{fig:The-overall-architecture}}
\end{figure*}

\begin{figure}[t]
\centering{}\centering{}\centering \resizebox{0.5\textwidth}{!}{
\begin{tabular}{@{}c}
\includegraphics[width=1\linewidth,height=1.4cm]{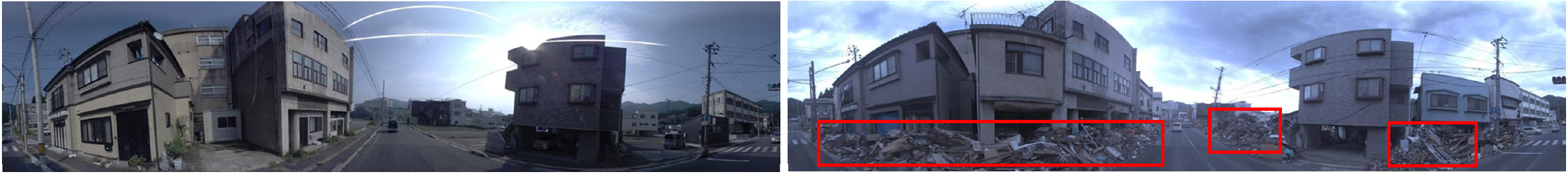}\tabularnewline
{\small{}(a) Images at $t_{0}$ and $t_{1}$}\tabularnewline
\includegraphics[width=1\linewidth,height=1.4cm]{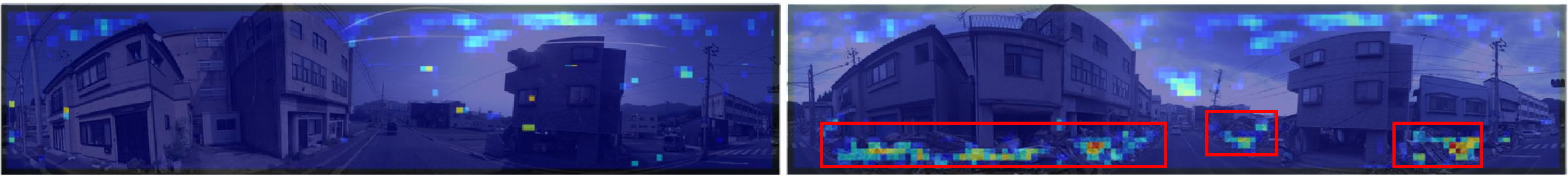}\tabularnewline
{\small{}(b) Feature maps at the same channel}\tabularnewline
\end{tabular}} \caption{Typical examples of paired feature maps. (a) Input images at different
times. (b) Paired feature maps at the 12-th channel, which are extracted
from the Conv4\_3 layer of the SCNN. We resize and overlay them on
original images for better visualization.\label{fig:Examples-of-extracted}}
\vspace{-2mm}

\end{figure}

\begin{table}[th]
\centering{}\centering{}\caption{The detailed configurations of our HPCFNet. In each row, the numbers
denote the corresponding channel number of the output feature map.
C, L, and B denote the layer of 3$\times3$ convolution, Leaky ReLU
and batch normalization. \label{tab:The-detailed-configurations}}
\doublerulesep=0.5pt \resizebox{0.42\textwidth}{!}{%
\begin{tabular}{ccccc}
\hline
\textbf{Location}  & \textbf{PCF}  & \textbf{RSA}  & \textbf{Conv Block}  & \textbf{MPFL}\tabularnewline
\hline
Conv1\_2  & 192  & -  & CLB, 64  & -\tabularnewline
Conv2\_2  & 384  & 384  & CLB, 128  & 64\tabularnewline
Conv3\_3  & 768  & 768  & CLB, 256  & 128\tabularnewline
Conv4\_3  & 1280  & 1280  & CLB, 512  & 256\tabularnewline
Conv5\_3  & 1024  & 1024  & CLB, 512  & 256\tabularnewline
\hline
\end{tabular}}
\end{table}

\section{Methodology\label{sec:Methodology}}

\subsection{Overall Architecture \label{subsec:Architecture-Overview}}

The overall architecture of the proposed HPCFNet is shown in Fig.
\ref{fig:The-overall-architecture}. For the SSCD task, the main steps
can be briefly described as follows. First, the paired images at different
times ($t_{0}$ and $t_{1}$) are fed into a Siamese Convolutional
Neural Network (SCNN) with the VGG-16 backbone~\cite{simonyan2014very}
to extract multi-level deep features. Then, feature maps from convolutional
layers (i.e., conv1\_2, conv2\_2, conv3\_3, conv4\_3 and conv5\_3)
are integrated by the proposed PCF modules to generate channel-wise
fused feature maps, hierarchically. Afterwards, the fused feature
maps are adjusted by the RSA modules. The MPFL strategy is inserted
to detect changes in a holistic-to-local manner. Finally, the change
map is predicted by combining the hierarchical fused features. Tab.
\ref{tab:The-detailed-configurations} shows the detailed configurations
of the proposed HPCFNet.

\subsection{Paired Channel Fusion\label{subsec:Paired-Channel-Fusion}}

To fuse feature maps, the most straightforward method is to concatenate
them in channel-wise. However, as a result, the characteristics of
the fused feature maps can't be well explored. Therefore, it is cumbersome
to detect change regions. By visualizing the channels of the SCNN
(shown in Fig. \ref{fig:Examples-of-extracted}), we observe that
the paired channels from the same layer in the two streams, can activate
most of the changed regions.
\begin{figure}[!t]
\begin{centering}
\includegraphics[width=0.9\linewidth,height=5.4cm]{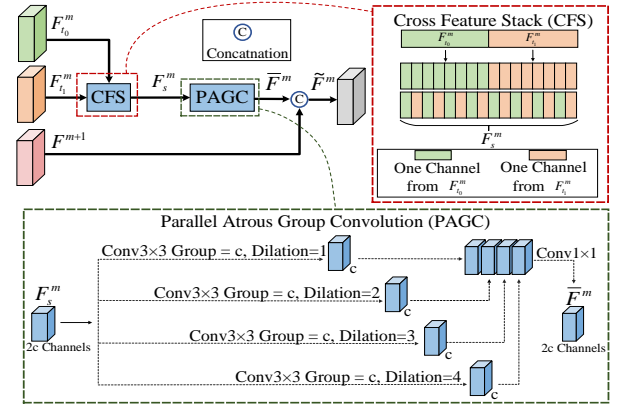}
\par\end{centering}
\caption{An illustration of the proposed PCF module. $F_{t_{0}}^{m}$ and $F_{t_{1}}^{m}$
are the $m$-th level feature maps extracted from images at $t_{0}$
and $t_{1}$, respectively.\label{fig:The-extracted-feature}}
\end{figure}

\begin{figure*}[!]
\centering \resizebox{0.9\textwidth}{!}{ %
\begin{tabular}{@{}c}
\includegraphics[width=0.8\linewidth]{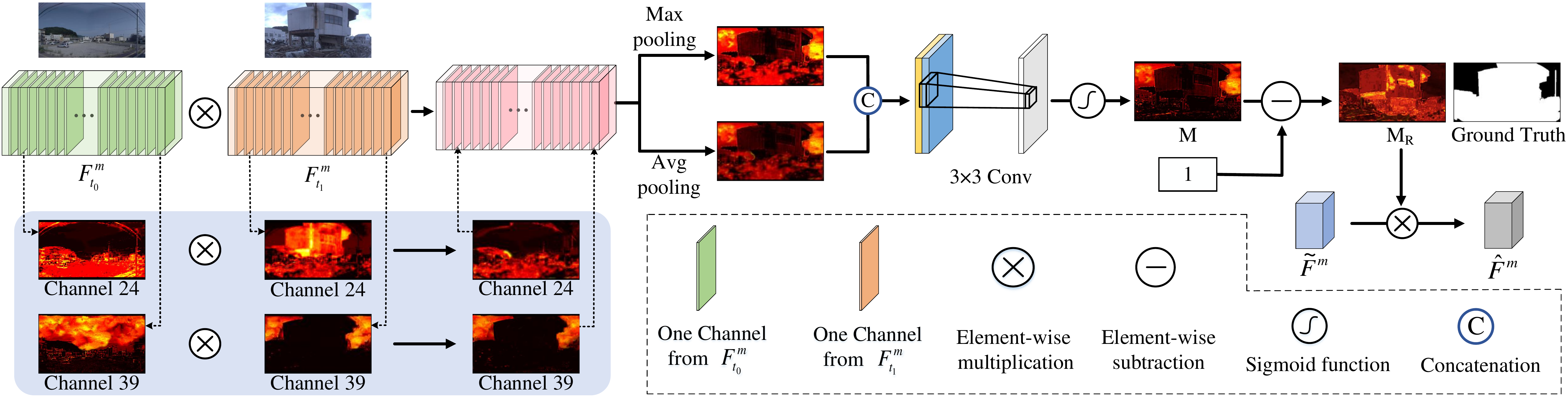}\tabularnewline
\end{tabular}} \caption{The detailed architecture of the RSA module. In order to show the
operation clearly, we here illustrate the changing process of channel
features for a definite input image pair.\label{fig:The-illustration-of-1}}
\vspace{-4mm}

\end{figure*}

Thus, paired channels extracted at the same level can contribute to
locating the changed regions. Inspired by this fact, we propose an
effective fusion method named Paired Channel Fusion (PCF) to incorporate
the information of paired channels. The overall architecture of the
PCF is shown in Fig. \ref{fig:The-extracted-feature}. More specifically,
at the $m$-th level, we first combine the same-level feature maps
(i.e., $\textbf{F}_{t_{0}}^{m}$ and $\textbf{F}_{t_{1}}^{m}$) by
the Cross Feature Stack (CFS) to make the channels interweave, resulting
in $\textbf{F}_{s}^{m}$. Motivated by Atrous Spatial Pyramid Pooling
(ASPP) module \cite{chen2017deeplab}, we propose a Parallel Atrous
Group Convolution (PAGC) module to fuse paired channels and capture
multi-scale feature representations. The PAGC module has four separate
group convolutions with kernel size of $3\times3$. For simplicity,
the group number of each group convolutional layer is set to $c$,
which is the channel number of the input $\textbf{F}_{t_{0}}^{m}$
or $\textbf{F}_{t_{1}}^{m}$. Thus, each group in the group convolutional
layer contains two channels: one channel from $\textbf{F}_{t_{0}}^{m}$,
and the corresponding channel from $\textbf{F}_{t_{1}}^{m}$. In addition,
we use dilated convolution to enlarge the receptive fields. The outputs
of four group convolutions are concatenated, then reduced through
a $1\times1$ convolutional layer to avoid too much computation. The
output of a PAGC (i.e., $\bar{\textbf{F}}^{m}$) is concatenated with
the processed feature map from the $(m+1)$-th level (i.e., $\textbf{F}^{m+1}$)
to produce the output of the PCF module (i.e., $\widetilde{\textbf{F}}^{m}$).

The PCF can be formulated as:
\begin{equation}
\small\bar{\textbf{F}}^{m}=g^{1\times1}([f_{d=1}^{3\times3}(\textbf{F}_{s}^{m}),f_{d=2}^{3\times3}(\textbf{F}_{s}^{m}),f_{d=3}^{3\times3}(\textbf{F}_{s}^{m}),f_{d=4}^{3\times3}(\textbf{F}_{s}^{m})]),
\end{equation}
\begin{equation}
\small\widetilde{\textbf{F}}^{m}=[\bar{\textbf{F}}^{m},\textbf{F}^{m+1}]
\end{equation}
where $f_{d=n}^{k\times k}$ denotes the atrous group convolution
which consists of (i) a convolutional layer with a group number $c$,
kernel size $k\times k$ and dilation $n$, (ii) a batch normalization~\cite{ioffe2015batch},
and (iii) a non-liner activation function. $[\cdot]$ denotes the
channel-wise concatenation. $g^{1\times1}$ is a $1\times1$ convolutional
layer.

\subsection{Reverse Spatial Attention\label{subsec:Reverse-Spatial-Attention}}

The fused feature map produced by the PCF module can coarsely locate
the changed regions, however, it lacks the capacity of highlighting
regional details. To solve this problem, we further introduce a Reverse
Spatial Attention (RSA) module, as shown in Fig. \ref{fig:The-illustration-of-1}.
We feed $\widetilde{\textbf{F}}^{m}$, $\textbf{F}_{t_{0}}^{m}$ and
$\textbf{F}_{t_{1}}^{m}$ into the RSA module to produce the weighted
fused feature map.

It is well known that, the Siamese network extracts features of two
different images in the same way. Hence the changeless regions in
paired images result in feature maps that contain the same semantic
information. Meanwhile, the changed regions result in different activations
on paired feature maps. In our RSA (Fig. \ref{fig:The-illustration-of-1}),
we first perform the element-wise multiplication ( $\textbf{F}_{t_{0}}^{m}$
and $\textbf{F}_{t_{1}}^{m}$) to enhance the same information. After
multiplication, each two corresponding channels will keep the changed
regions non-activated. This is due to changed regions will lead to
large different activation values in corresponding channels. From
Fig. \ref{fig:The-illustration-of-1}, one can observe that whether
in channel 24 or channel 39, the changed area is non-activated after
multiplication. Due to the limitation of space, we only show the changing
process of channel 24 and channel 39. The other channels have a similar
trend. Therefore, after pooling along the channel axis, the changed
regions are remain non-activated. Specifically, an average-pooling
and a max-pooling are separately executed along the channel axis to
capture the statistical properties of feature maps~\cite{woo2018cbam}.
The resulting feature maps are concatenated and fed into a $3\times3$
convolutional layer to generate a spatial attention mask $\textbf{M}\in R^{h\times w\times1}$.
Because the multiplied features emphasize the same representations
of two feature maps. The generated $\textbf{M}$ is capable of highlighting
the changeless regions. However, for change detection, we prefer to
highlight the changed regions. Thus, the reversed mask is more effective
for the SSCD task. Formally, the mask $\textbf{M}_{R}$ can be computed
by:
\begin{equation}
\small\textbf{M}_{R}=\textbf{1}-\sigma(\textbf{W}^{T}[AP(\textbf{F}_{t_{0}}^{m}\odot\textbf{F}_{t_{1}}^{m}),MP(\textbf{F}_{t_{0}}^{m}\odot\textbf{F}_{t_{1}}^{m})]+\textbf{b}),
\end{equation}
where $\sigma$ denotes a sigmoid function. $\textbf{W}$ and $\textbf{b}$
denote learn-able parameters. $AP(\cdot)$ and $MP(\cdot)$ are the
average-pooling and max-pooling, respectively. $\odot$ denotes the
element-wise multiplication. Based on the RSA mask, the attentive
features can be expressed as:
\begin{equation}
\small\hat{\textbf{F}}_{i,j,k}^{m}=\textbf{M}_{R}\odot\widetilde{\textbf{F}}_{i,j,k}^{m},
\end{equation}
where $(i,j,k)$ denotes the width, height and channel index of the
feature maps. Note that the attention mask $\textbf{M}_{R}$ is duplicated
$k$ times to weight every channel.

\subsection{Multi-Part Feature Learning\label{subsec:Multi-part-Feature-Learning}}

\begin{figure}[t]
\begin{centering}
\includegraphics[width=1\linewidth]{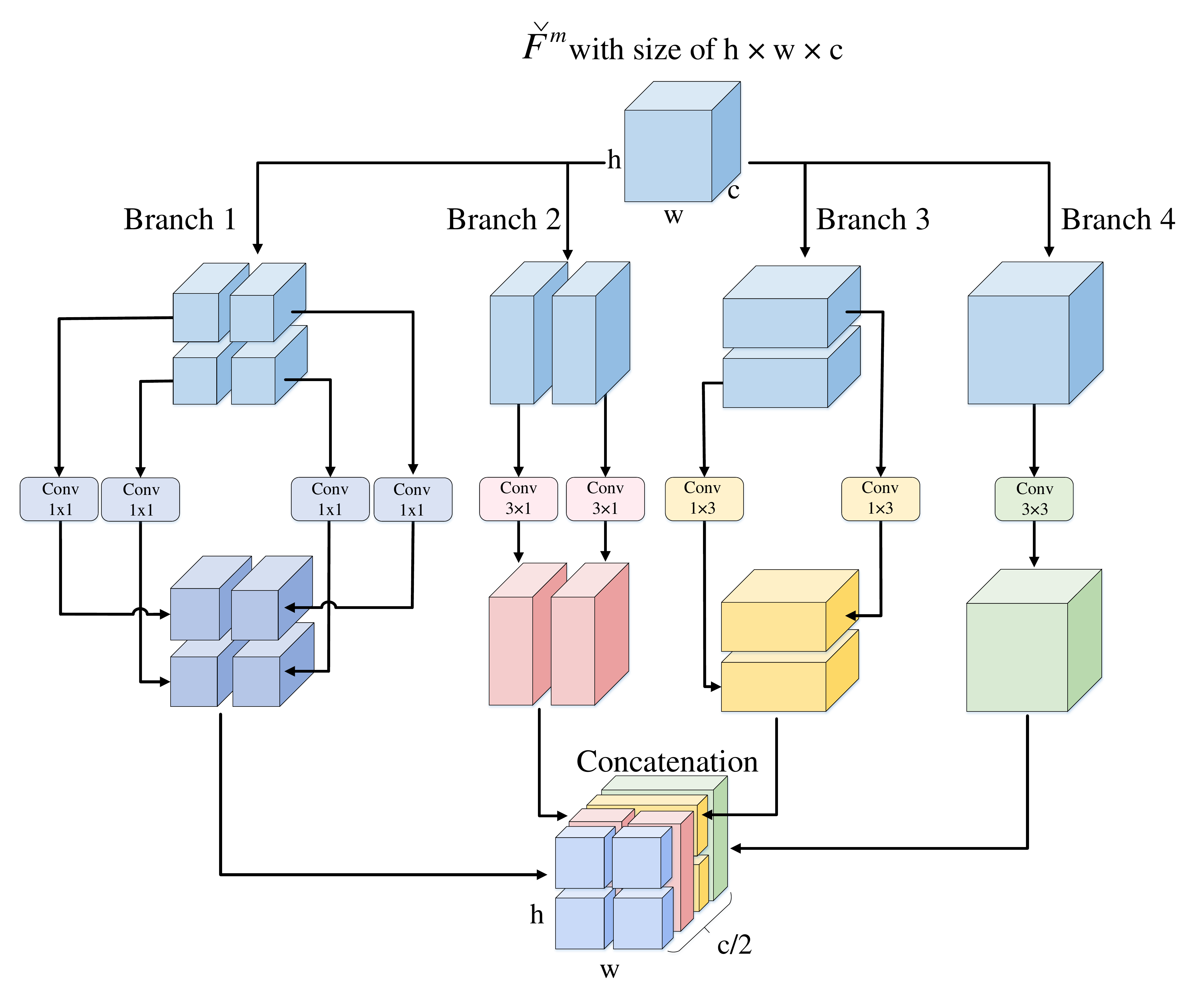}
\par\end{centering}
\caption{An illustration of the Multi-Part Feature Learning (MPFL) strategy.
The feature map is fed into 4 independent branches. \label{fig:The-illustration-of}}
\vspace{-2mm}

\end{figure}

As mentioned in Section I, both the locations and the scales of changed
regions are unbalanced in the dataset. To address this problem, we
propose a novel MPFL strategy which detects changes from global to
local. As shown in Fig. \ref{fig:Examples-from-three}, we observe
that the diversity can be divided into four cases. Therefore we design
four corresponding partitions for above cases, which are presented
in Fig. \ref{fig:The-illustration-of}. More specifically, given the
attentive feature $\hat{\textbf{F}}^{m}$, a $1\times1$ convolutional
layer is first performed to reduce the channel dimension, resulting
in $\check{\textbf{F}}^{m}$. Then, we partition the features as follows:
\ding{172} dividing the original feature $\check{\textbf{F}}^{m}$
into 4 feature blocks (Branch 1) along the height and width where
each feature block has the size of $(h/2)\times(w/2)\times c$; \ding{173}
dividing the $\check{\textbf{F}}^{m}$ into $2$ feature blocks (Branch
2) only along the width axis where each feature block has the size
of $h\times(w/2)\times c$; \ding{174} dividing the $\check{\textbf{F}}^{m}$
into $2$ feature blocks (Branch 3) only along the height axis where
each feature block has the size of $(h/2)\times w\times c$; \ding{175}
The final branch (Branch 4) is the original feature $\check{\textbf{F}}^{m}$.
For each branch, we set specific kernel sizes to adapt to different
scales of feature blocks, i.e., $1\times1$ in Branch 1, $3\times1$
in Branch 2, $1\times3$ in Branch 3 and $3\times3$ in Branch 4.
Note that in each branch, the convolution of each feature block is
independent. Hence MPFL can adaptively learn global and local features
with appropriate receptive fields. Finally, we concatenate all the
feature maps of different branches. In the proposed MPFL strategy,
the four branches utilize different spatial partition approaches and
adaptive convolutional layers, thus they can capture the discriminative
characteristics from different spatial regions. Based on the MPFL
strategy, the performance of SSCD can be significantly improved as
shown in Section V. D.

\subsection{Network Training \label{subsec:Network-Training}}

Given the training dataset as $\left\{ (X_{t_{0}}^{n},X_{t_{1}}^{n}),Y^{n}\right\} _{n=1}^{N}$,
where $(X_{t_{0}}^{n},X_{t_{1}}^{n})$ is the paired image, $Y^{n}$
is the ground-truth change map, and $N$ is the total number of training
examples. $Y_{l}^{n}$ denotes the $l$-th pixel of $Y^{n}$. Without
loss of generality, we subsequently drop the superscript $n$ and
consider each pixel for the network training.

The softmax cross-entropy loss function is effective for most image
pairs which include large change areas. However, for a typical natural
image, the class distribution of changed/non-changed pixels is heavily
imbalanced. To relieve the class-imbalance problem, we adopt the weighted
cross-entropy loss function to train our model. Formally, the loss
function can be calculated by:
\begin{equation}
\mathcal{L}=-\frac{1}{L}\sum_{l=1}^{L}\sum_{j=0}^{1}w_{j}\text{log~Pr}(Y_{l}=j|(X_{t_{0}}^{n},X_{t_{1}}^{n});\theta),
\end{equation}
where $w_{j}$ is a weight for class $j$, $\text{Pr}(Y_{l}=j|X;\theta)$
is the probability that measures how likely the pixel belongs to the
$j$-th class. For the weights, we adopt:
\begin{equation}
w_{0}=1-\frac{n_{T}}{n_{T}+n_{F}},w_{1}=\frac{n_{T}}{n_{T}+n_{F}},
\end{equation}
where $n_{T}$ denotes the number of pixels which are set in true
class, and $n_{F}$ denotes that of pixels in the other class. The
$n_{T}+n_{F}$ is the total number of pixels. The above loss function
is continuously differentiable, so we can use the standard Stochastic
Gradient Descent (SGD) method \cite{deng2009imagenet} to obtain the
optimal parameters.

\section{Experimental Setups\label{sec:Experimental-Setups}}

In order to demonstrate the effectiveness of the proposed method,
we evaluate it on three public benchmark datasets, i.e., PCD \cite{sakurada2015change},
VL-CMU-CD \cite{alcantarilla2018street} and CDnet2014 \cite{wang2014cdnet}.
We first introduce details of the three publicly available datasets.
Then, we present the evaluation metrics and implementation details.
Afterwards, we compare our results against other state-of-the-art
methods. Finally, we conduct ablation experiments to demonstrate the
effects of different modules.

\subsection{Street Scene Change Detection Datasets}

The \textbf{PCD} dataset \cite{sakurada2015change} consists of two
subsets, i.e., ``GSV'' and ``TSUNAMI''. Each subset consists of
100 panoramic image pairs and the hand-labeled change masks. The camera
viewpoints of every image pair are different. In detail, the GSV dataset
consists of 100 panoramic image pairs of Google Street View, and the
TSUNAMI dataset consists of 100 panoramic image pairs of scenes after
a tsunami.

The \textbf{VL-CMU-CD} dataset \cite{alcantarilla2018street} is a
street-view change detection dataset with a long time span. It contains
151 image sequences for change detection. It can generate a total
of 1362 image pairs, each pair of them can be provided with a ground
truth with labeling mask for five classes. The split sets contain
933 training image pairs and 429 test image pairs.

The \textbf{CDnet2014} dataset \cite{wang2014cdnet} contains 53 video
sequences and is created for Foreground Object Extraction
(FOE) and Moving Object Segmentation (MOS). Each video sequence contains
frames from 600 to 7999 with resolutions varying from $320\times240$
to $720\times576$. The dataset includes changes to illumination,
shadows, camera viewpoint and background movement. In
general, foreground detection can be regarded as change detection
based on multi-frame sequences. Hence, we strictly follow CosimNet
\cite{guo2018learning} to implement our SSCD method on CDnet2014.
Please refer to Section \ref{subsec:Implementation-Details} for implementation
details.

\subsection{Evaluation Metrics}

Following previous works, we evaluate the performance based on the
widely-used F-Score. This metric is a weighted harmonic mean of Precision
and Recall. It ranges from 0 to 1, and higher value indicates better
performance. More specifically, it is computed as follows:
\begin{equation}
\small\text{F-Score}=\frac{(\beta^{2}+1)\times Precision\times Recall}{\beta^{2}\times Precision+Recall},
\end{equation}
where $\beta$ is a balanced hyper-parameter. As previous works suggested,
we set $\beta$ to 1 to weight $Precision$ and $Recall$ equally.
Given true positive (TP), false positive (FP), false negative (FN),
true negative (TN), F-Score can be deduced by the four fundamental
metrics. Besides, $Precision=\frac{TP}{TP+FP},Recall=\frac{TP}{TP+FN}$.

\subsection{Implementation Details \label{subsec:Implementation-Details}}

We implement our proposed framework with PyTorch\cite{paszke2017automatic}
in Python 3.6. The HPCFNet was trained and tested on a workstation
with 8 NVIDIA RTX 2080Ti GPUs (each with 11G memory) and two E5-2620
CPUs.
\begin{figure}
\centering{}\centering{}\centering \resizebox{0.47\textwidth}{!}{
\includegraphics[width=1\linewidth]{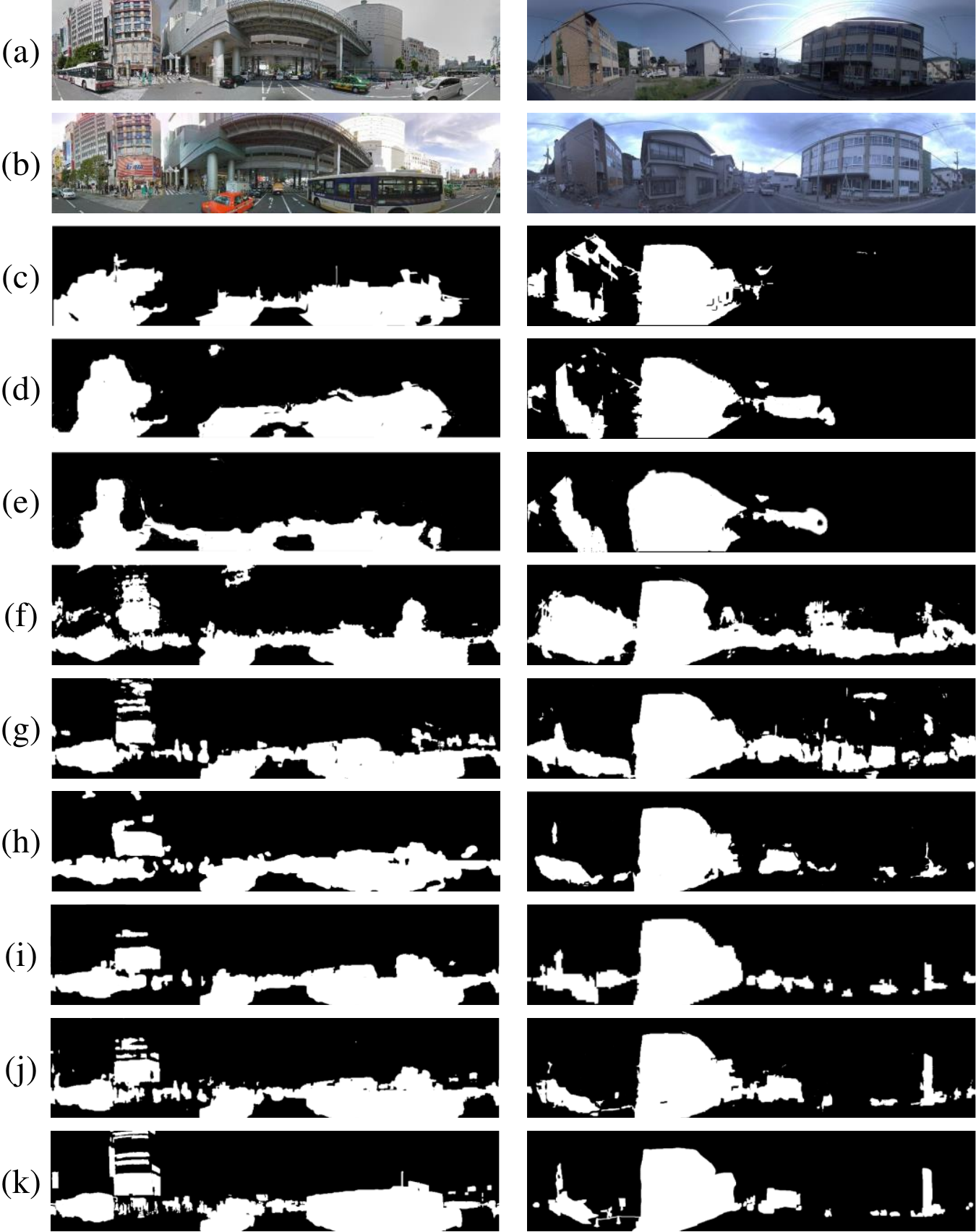}} 
\caption{Visual comparison with different methods on the PCD dataset. (a) Image
at $t_{0}$, (b) Image at $t_{1}$, (c) CNN-feat~\cite{sakurada2015change},
(d) DAISY \cite{tola2009daisy}, (e) DASC \cite{kim2015dasc}, (f)
DN \cite{alcantarilla2018street}, (g) CosimNet \cite{guo2018learning},
(h) DOF-CDNet \cite{sakurada2017dense}, (i) CSCDNet \cite{sakurada2018weakly},
(j) Ours, (k) Ground Truth. \label{fig:The-comparisons-among}}
\vspace{-4mm}
\end{figure}

\textbf{Data Preprocessing}. (i) For the PCD dataset, it contains
200 panoramic image pairs, each of which has a 224$\times$1024 resolution.
To train our model, we crop the original image by sliding 56 pixels
in width, hence each panoramic image will generate 15 patches with
a $224\times224$ resolution. After the plane rotation and mirror,
24000 image pairs are generated in total. Following previous works,
we adopt 5-fold cross-validation for model training and testing.

(ii) For the VL-CMU-CD dataset, we also perform data augmentation
with the plane rotation and mirror, as the PCD dataset. During the
training procedure, we resize the paired images to 512$\times$512.
We follow the official split \cite{alcantarilla2018street}, using
933/429 for training and testing. Note that we reduce the multi-class
labeling mask to a binary change map, because we focus on change regions
instead of class information.

(iii) CDnet2014 consists of 53 moving video sequences but not each
image is labeled with a pixel-wise ground-truth. In terms of the evaluations
on CDnet2014, we strictly follow previous works, especially the CosimNet
\cite{guo2018learning}, to conduct experiments including the selection
of image pairs, the split of training/validation set and the process
of testing. Following the CosimNet, we selected the background images
(i.e., without any foreground objects) as the reference image at time
$t_{0}$ and others as the query images at $t_{1}$. The CDnet2014
includes 53 video sequences in different scenarios. We used the annotated
frames to built a total of 91595 image pairs, which consist of a training
set and a validation set with 73276 pairs and 18319 for each. As suggested
by the CDnet2014 benchmark, we evaluate our model on the test set
of CDnet2014. Specifically, the image without any change-inducing
object is selected as one of the comparison images at time $t_{0}$,
and others as the image at time $t_{1}$. All images are resized to
512$\times$512 during training and validating.

\textbf{Parameter Settings}. For the experiment, the proposed framework
is initialized from the VGG-16 model\cite{simonyan2014very}, which
was pre-trained on the ImageNet dataset \cite{deng2009imagenet}.
The weights of other new layers are initialized by the ``xavier''
method \cite{glorot2010understanding}. In the training phase, we
use the the standard Stochastic Gradient Descent (SGD) \cite{krizhevsky2012imagenet}
optimizer. We set the batch size to 8 and the base learning rate to
$1e^{-2}$. The momentum is set to 0.95 and weight decay to $1.25e^{-4}$.
After 200 epochs, the training procedure converged. We will release
the source codes upon acceptance of this work.

\section{Experimental Results\label{sec:Experimental-Results}}

\subsection{Evaluation on The PCD Dataset}

Tab. \ref{tab:PCD} shows the quantitative results on the PCD dataset.
It can be seen that the F-Score of our framework is higher than other
baseline methods. More specifically, our framework achieves a remarkable
F-Score (0.776) on the GSV dataset. There is about 4\% improvement
to the second-best method, i.e., CSCDNet \cite{sakurada2018weakly}.
Besides, our framework also achieves 0.868 on the TSUNAMI dataset,
about 1\% improvement compared to the CSCDNet.

Note that both the CDNet and DOF-CDNet \cite{sakurada2017dense} adopt
the U-Net \cite{ronneberger2015u} for change detection. Due to the
limited representation ability, they show much worse results than
our method. The CosimNet \cite{guo2018learning} also adopts a Siamese
network, which is based on a more powerful segmentation model, i.e.,
the pre-trained Deeplabv2~\cite{chen2017deeplab}. However, our method
also shows better results than the CosimNet. Several qualitative results
are shown in Fig. \ref{fig:The-comparisons-among}. It can be seen
that our framework is able to capture small as well as large change
regions (e.g., large buildings, big cars, small pedestrians, slender
poles, imperceptible advertising boards).
\begin{table}[!th]
\centering{}\centering{}\caption{Quantitative results on the PCD dataset. The best two results are
in bold and underline. \label{tab:PCD}}
\centering{}\doublerulesep=0.5pt \resizebox{0.48\textwidth}{!}{
\begin{tabular}{cccc}
\hline
\multirow{2}{*}{Methods} & \multirow{2}{*}{CNN Backbone} & \multicolumn{2}{c}{F-Score}\tabularnewline
\cline{3-4} \cline{4-4}
 &  & GSV  & TSUNAMI \tabularnewline
\hline
Dense SIFT \cite{lowe2004distinctive}  & -  & 0.528  & 0.649\tabularnewline
DAISY \cite{tola2009daisy}  & -  & 0.377  & 0.529\tabularnewline
DASC \cite{kim2015dasc}  & -  & 0.409  & 0.622\tabularnewline
CNN-feat \cite{sakurada2015change}  & AlexNet  & 0.639  & 0.724\tabularnewline
DN \cite{alcantarilla2018street}  & DeconvNet  & 0.614  & 0.774\tabularnewline
CosimNet \cite{guo2018learning}  & Deeplabv2  & 0.692  & 0.806\tabularnewline
CDNet \cite{sakurada2017dense}  & U-Net  & 0.693  & 0.838\tabularnewline
DOF-CDNet \cite{sakurada2017dense}  & U-Net  & 0.703  & 0.838\tabularnewline
CSCDNet \cite{sakurada2018weakly}  & Res-18  & \uline{0.738}  & \uline{0.859}\tabularnewline
\hline
HPCFNet (Ours)  & VGG-16  & \textbf{0.776}  & \textbf{0.868} \tabularnewline
\hline
\end{tabular}}
\end{table}

\begin{figure*}
\centering{}\centering{}\centering \resizebox{1.0\textwidth}{!}{
\includegraphics[width=1\linewidth,height=8.5cm]{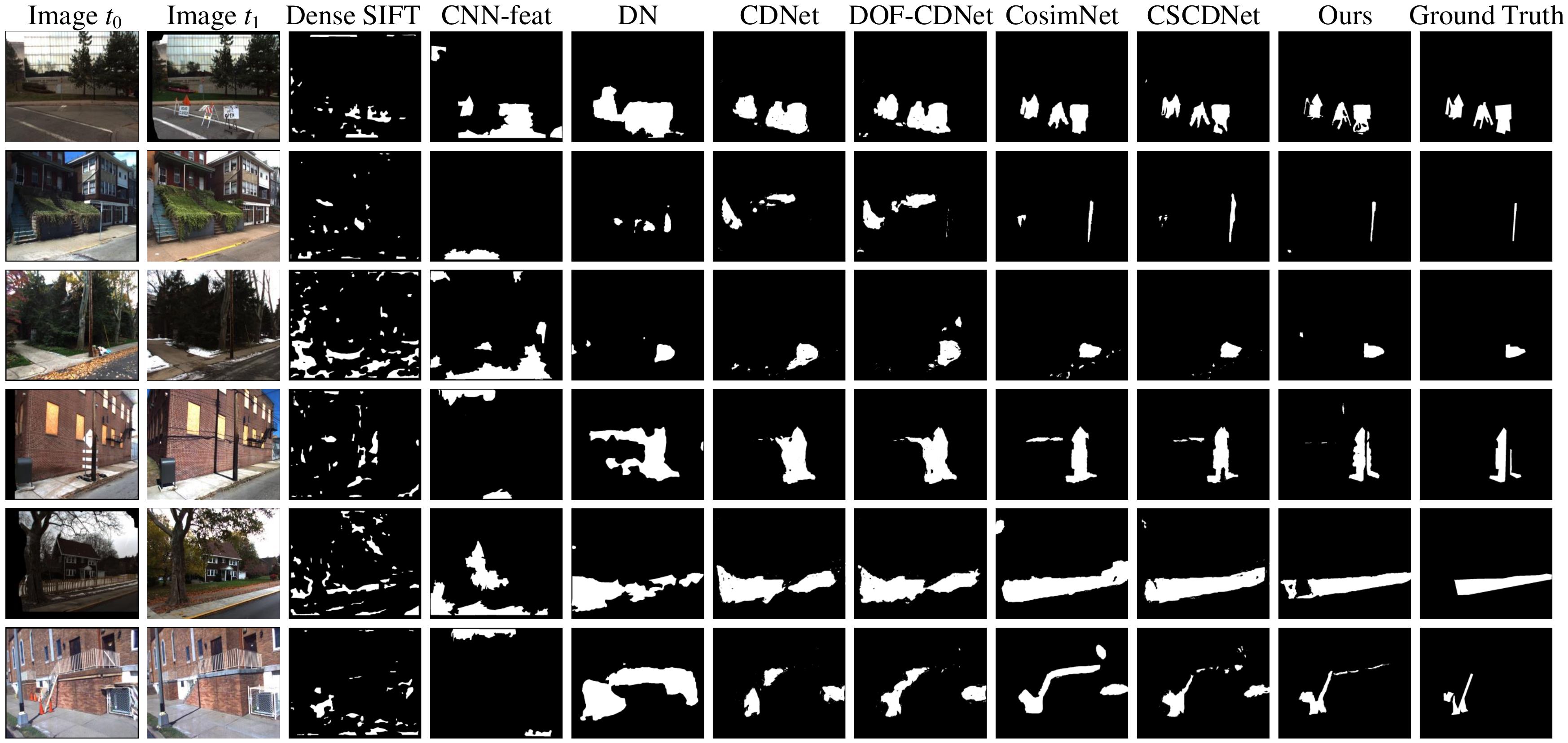}
} \caption{Visual comparisons on the VL-CMU-CD dataset. Our method is good at
locating and resolving details. \label{fig:Visual-comparisons-on}}
\vspace{-4mm}

\end{figure*}

\subsection{Evaluation on The VL-CMU-CD Dataset}

Tab. \ref{tab:CMU} shows the performance of compared methods on the
VL-CMU-CD dataset. In terms of F-Score, our framework achieves an
accuracy of 0.752, and outperforms the second best CSCDNet \cite{sakurada2018weakly}
by 4.2\%. Fig. \ref{fig:Visual-comparisons-on} shows several typical
visual results on the VL-CMU-CD dataset. It can be seen that the proposed
framework performs well and is visibly better than other methods.
Especially, our framework is good at locating and resolving details,
e.g, the 1-2 rows of Fig.~\ref{fig:Visual-comparisons-on}. And it
is more accurate on the boundary of changed regions, e.g., the 4 row
of Fig.~\ref{fig:Visual-comparisons-on}.

\begin{table}[!th]
\centering{}\centering{}\caption{Quantitative results on the VL-CMU-CD dataset. The best two results
are in bold and underline. \label{tab:CMU}}
\centering{}\doublerulesep=0.5pt \resizebox{0.48\textwidth}{!}{%
\begin{tabular}{cccc}
\hline
Methods  & CNN Backbone  & Publication  & F-Score\tabularnewline
\hline
Dense SIFT \cite{lowe2004distinctive}  & -  & IJCV 2004  & 0.243\tabularnewline
DAISY \cite{tola2009daisy}  & -  & TPAMI 2009  & 0.181\tabularnewline
DASC \cite{kim2015dasc}  & -  & CVPR 2015  & 0.234\tabularnewline
CNN-feat \cite{sakurada2015change}  & AlexNet  & BMVC 2015  & 0.403\tabularnewline
DN \cite{alcantarilla2018street}  & DeconvNet  & AR 2018  & 0.582\tabularnewline
CDNet \cite{sakurada2017dense}  & U-Net  & Arxiv 2017  & 0.685\tabularnewline
DOF-CDNet \cite{sakurada2017dense}  & U-Net  & Arxiv 2017  & 0.688\tabularnewline
CosimNet \cite{guo2018learning}  & Deeplabv2  & Arxiv 2018  & 0.706\tabularnewline
CSCDNet \cite{sakurada2018weakly}  & Res-18  & Arxiv 2018  & \uline{0.710}\tabularnewline
\hline
HPCFNet (Ours)  & VGG-16  &  & \textbf{0.752}\tabularnewline
\hline
\end{tabular}}
\end{table}

\begin{figure}[!t]
\centering{}\centering{}\centering \resizebox{0.48\textwidth}{!}{
\begin{tabular}{c}
\includegraphics{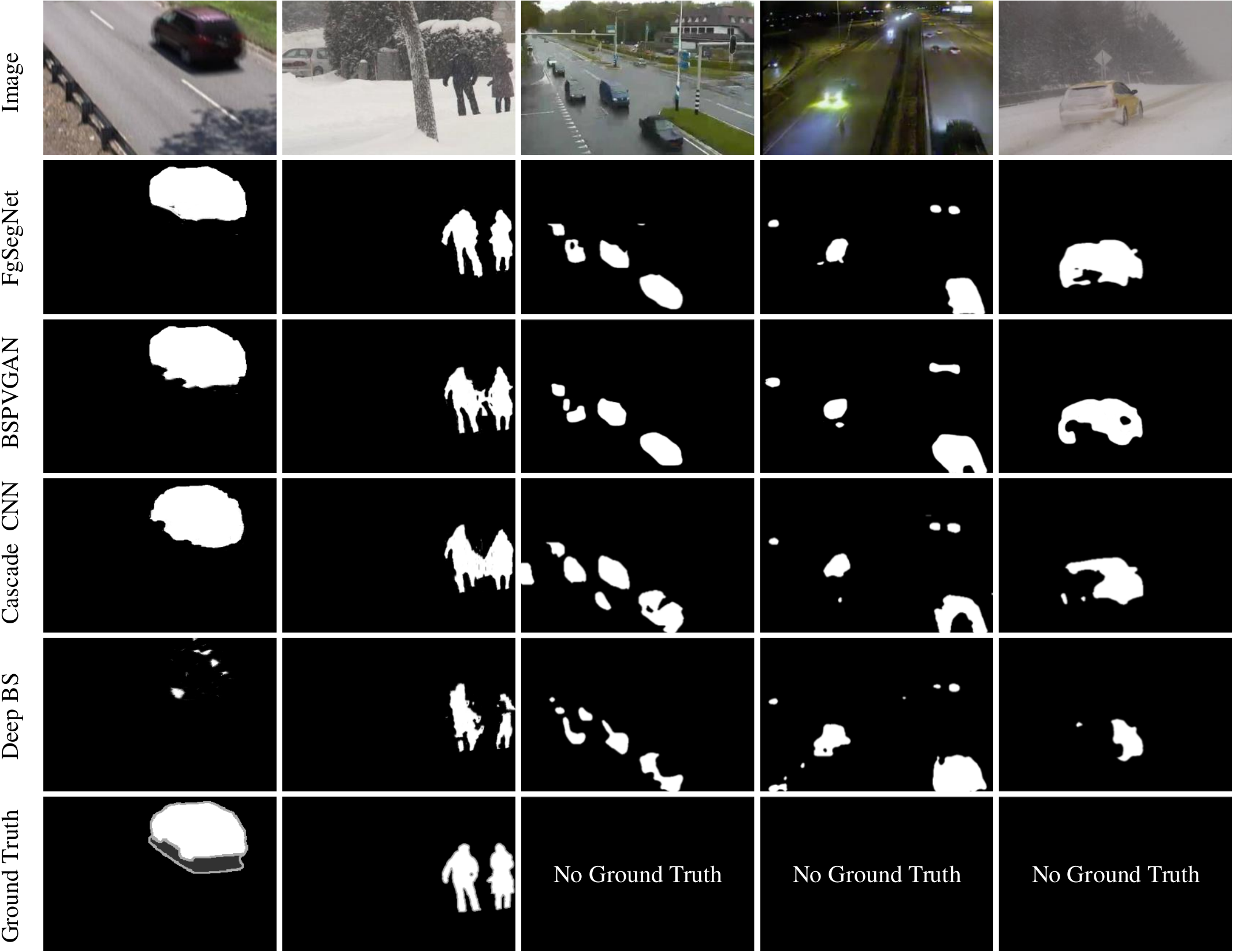}\tabularnewline
\tabularnewline
\includegraphics{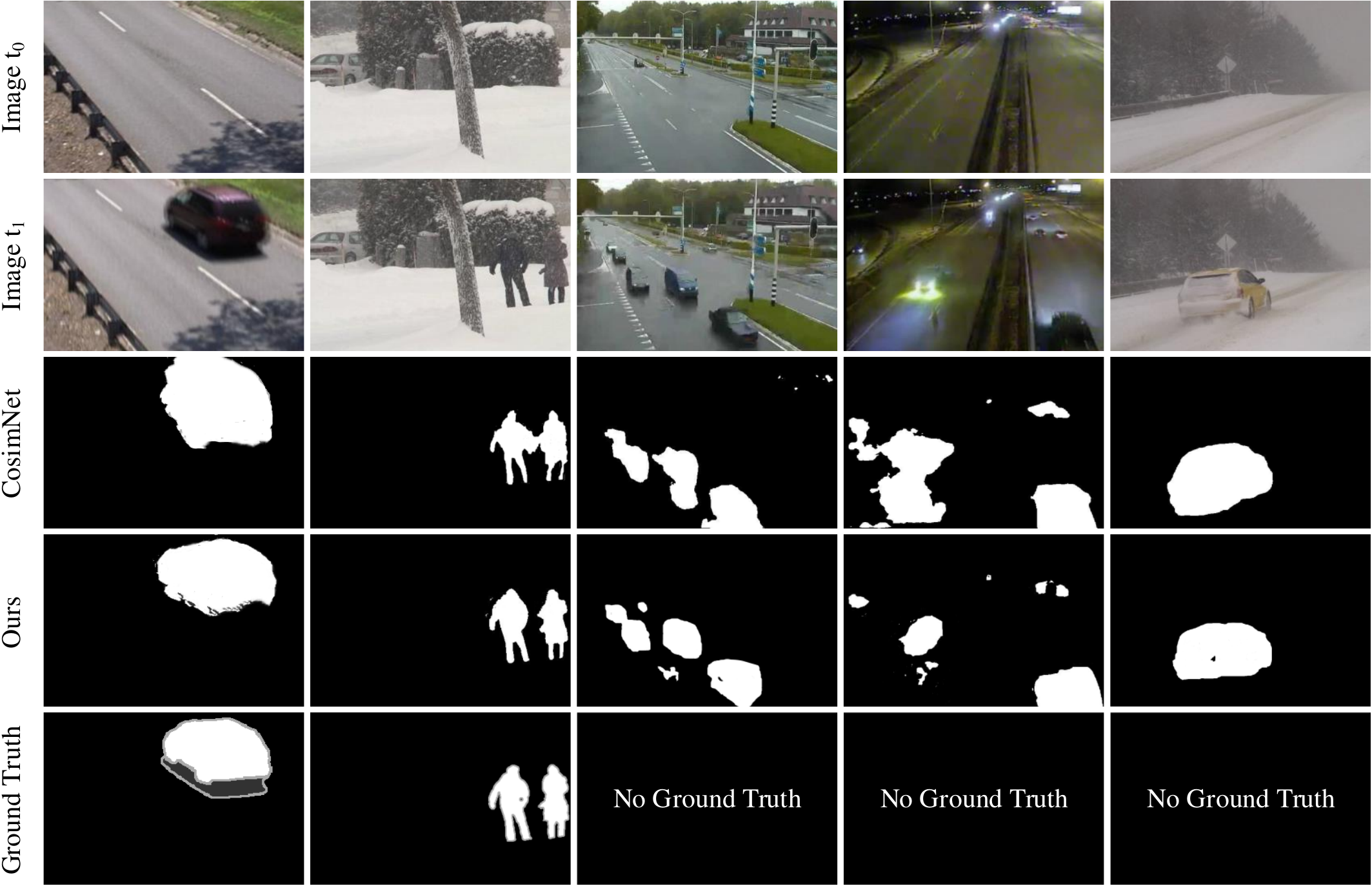}\tabularnewline
\tabularnewline
\end{tabular}} \caption{Qualitative results on the CDnet2014 dataset. Top sub-figure: Examples
with top-ranked FOE and MOS methods. Bottom sub-figure: Examples with
SSCD methods. \label{fig:Qualitative-results-on}}
\end{figure}

\subsection{Evaluation on The CDnet2014 Dataset}

\begin{figure*}[!th]
\begin{centering}
\includegraphics[width=1\linewidth,height=5.2cm]{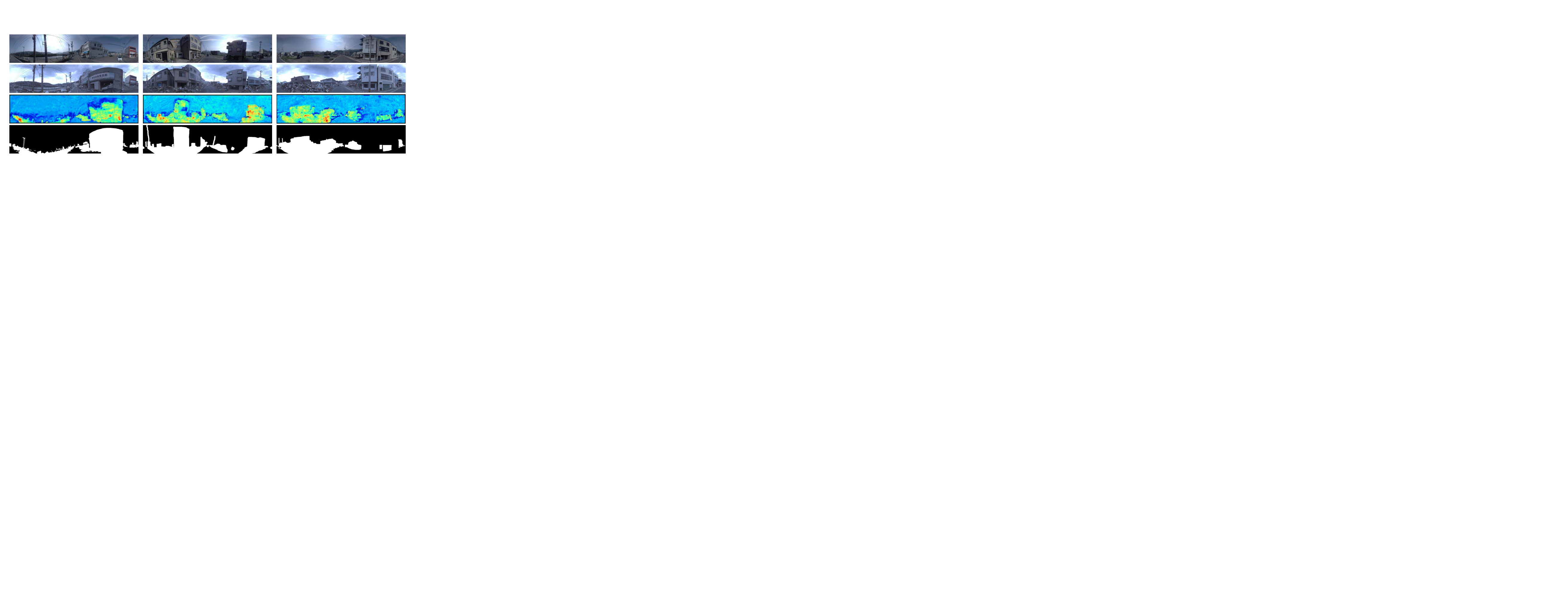}
\par\end{centering}
\caption{Illustration of Reverse Spatial Attention masks. The first and second
rows show the images at $t_{0}$ and $t_{1}$, respectively. The third
row shows our RSA masks, and the last row indicates the ground-truths.
One can see that the highlighted regions by our RSA module are coarsely
consistent to the changed regions. \label{fig:A-comparison-on}}
\end{figure*}

Tab. \ref{tab:CDNet} shows the performance of different methods on
the CDnet2014 dataset. The methods at the last two rows are the SSCD
methods. Note that the CDnet2014 is essentially created
for Foreground Object Extraction (FOE) and Moving Object Segmentation
(MOS). In other rows, we also show several top-ranked methods for
FOE and MOS. On the tasks of FOE and MOS, researchers mainly devote
to segment the foreground object for each frame in video sequence.
However, HPCFNet is a method for SSCD aiming to detect the scene change
rather than foreground moving objects only. Strictly speaking, CDnet2014
is not a standard dataset for SSCD. Thus, the evaluation performance
of HPCFNet is worse than several methods for FOE and MOS. Even though,
our framework achieves 0.863 in terms of F-Score, which is very competitive
to several methods for FOE and MOS. Besides, two key reasons motivate
us to perform experiments on CDnet2014. (a) To the best of our knowledge,
CosimNet is the only SSCD method that is evaluated on CDnet2014. To
make a fair comparison with CosimNet, we follow it to conduct experiment
on CDnet2014. (b) We want to demonstrate that our method is feasible
for FOE and MOS on CDnet2014.

Meanwhile, our proposed framework surpasses the CosimNet \cite{guo2018learning},
which shows supreme performance for the SSCD task. The results are
reported by the online evaluation server\footnote{http://jacarini.dinf.usherbrooke.ca/results2014/662/.}.
The qualitative results are shown in Fig. \ref{fig:Qualitative-results-on}.
To better verify the generalization, we randomly choose some results
from the validating data which are provided ground
truths, we also randomly choose some examples from the testing data
which have no ground truth. Our method shows outstanding results for
images acquired in varying conditions (e.g., hard light, dim light
and bad weather).
\begin{table}[!th]
\centering{}\centering{}\caption{Quantitative results on the CDnet2014 dataset. The best two results
are in bold and underline. \label{tab:CDNet}}
\centering{}\doublerulesep=0.5pt \resizebox{0.38\textwidth}{!}{%
\begin{tabular}{ccc}
\hline
Task  & Methods  & F-Score \tabularnewline
\hline
\multirow{8}{*}{FOE / MOS } & SuBSENSE \cite{st2014subsense}  & 0.741 \tabularnewline
 & DeepBS \cite{babaee2018deep}  & 0.759 \tabularnewline
 & SemanticBGS \cite{braham2017semantic}  & 0.789 \tabularnewline
 & IUTIS-3 \cite{bianco2017combination}  & 0.751 \tabularnewline
 & CP3-online \cite{liang2015co}  & 0.792 \tabularnewline
 & Cascade CNN \cite{wang2017interactive}  & 0.921 \tabularnewline
 & FgSegNet V2 \cite{lim2020learning}  & \textbf{0.985} \tabularnewline
 & BSPVGAN \cite{zheng2020novel}  & \uline{0.950 }\tabularnewline
\hline
\multirow{2}{*}{SSCD} & CosimNet \cite{guo2018learning}  & 0.859 \tabularnewline
 & HPCFNet (Ours)  & 0.863 \tabularnewline
\hline
\end{tabular}} \vspace{-4mm}
\end{table}

\subsection{Ablation Studies\label{sec:Ablation-Study}}

To demonstrate the benefits of our proposed modules, we conduct additional
experiments by adding the modules progressively. Due to the limitation
of space, we only exhibit the results on the GSV and TSUNAMI datasets.
The other datasets have a similar performance trend. Note that, all
the experiments below are based on the VGG-16 backbone.

\textbf{Effects of PCF modules.} To verify the PCF module, a series
of experiments are performed, including the framework with concatenation
fusion or PCF fusion under different fusion architectures (i.e., early
fusion, late fusion and dense fusion). For example, ``Early Fusion
(Concat)'' denotes the early fusion with channel-wise concatenation.

As reported in Tab. \ref{tab:PCF}, with the PCF module and a dense
fusion, the model improves the performance by a large margin. We here
observe that the models with PCF consistently outperform than that
of only using the channel concatenation operators. This confirms the
effectiveness of our PCF. The main reason is that the fusion method
adds additional available information, which enriches the feature
representation. Only with one PCF module, our framework (f) achieves
the outstanding results on GSV$/$TSUNAMI datasets. It has achieved
comparable performance to most deep learning methods. These facts
further prove the effectiveness of our proposed method.

\begin{table}[!th]
\caption{Performance comparison on the PCF module. The best results are in
bold. \label{tab:PCF}}

\centering{}\centering{}\centering{}\doublerulesep=0.5pt \resizebox{0.48\textwidth}{!}{%
\begin{tabular}{ccccc}
\hline
\multirow{2}{*}{} & \multirow{2}{*}{Methods} & \multicolumn{3}{c}{F-Score}\tabularnewline
\cline{3-5} \cline{4-5} \cline{5-5}
 &  & GSV  &  & TSUNAMI\tabularnewline
\hline
(a)  & Early Fusion (Concat)  & 0.694  &  & 0.783\tabularnewline
(b)  & Early Fusion (PCF)  & 0.712  &  & 0.792\tabularnewline
(c)  & Late Fusion (Concat)  & 0.718  &  & 0.801\tabularnewline
(d)  & Late Fusion (PCF)  & 0.726  &  & 0.823\tabularnewline
(e)  & Dense Fusion (Concat)  & 0.728  &  & 0.824\tabularnewline
(f)  & Dense Fusion (PCF)  & \textbf{0.755}  &  & \textbf{0.857}\tabularnewline
\hline
\end{tabular}}
\end{table}

\textbf{Effects of the Cross Feature Stack (CFS). }To demonstrate
the effectiveness of the CFS module, we remove the CFS in PCF and
replace with the simple concatenation for comparison. Sufficient experiments
are conducted on the basis of three fusion structures: early fusion,
late fusion and dense fusion. The comparisons between CFS with concatenation
are shown in Table \ref{tab:CFS}. We can see that whichever structure
is chosen, CFS enables the boost of performance on both two datasets
(GSV and TSUNAMI). This fact indicates that CFS is beneficial for
Paired Channel Fusion.

\begin{table}[!th]
\caption{Performance comparison on the CFS module. The best results are in
bold. \label{tab:CFS}}

\centering{}\centering{}\centering{}\doublerulesep=0.5pt \resizebox{0.48\textwidth}{!}{%
\begin{tabular}{cccccc}
\hline
\multirow{2}{*}{Method} & \multirow{2}{*}{Fusion Structure} & \multirow{2}{*}{Fusion Manner} & \multicolumn{3}{c}{F-Score}\tabularnewline
\cline{4-6} \cline{5-6} \cline{6-6}
 &  &  & GSV  &  & TSUNAMI\tabularnewline
\hline
\multirow{6}{*}{PCF} & \multirow{2}{*}{Early} & concatenation  & 0.701  &  & 0.787\tabularnewline
 &  & CFS  & 0.712  &  & 0.792\tabularnewline
\cline{2-6} \cline{3-6} \cline{4-6} \cline{5-6} \cline{6-6}
 & \multirow{2}{*}{Late} & concatenation  & 0.720  &  & 0.813\tabularnewline
 &  & CFS  & 0.726  &  & 0.823\tabularnewline
\cline{2-6} \cline{3-6} \cline{4-6} \cline{5-6} \cline{6-6}
 & \multirow{2}{*}{Dense} & concatenation  & 0.737  &  & 0.831\tabularnewline
 &  & CFS  & \textbf{0.755}  &  & \textbf{0.857}\tabularnewline
\hline
\end{tabular}}
\end{table}

\begin{table}[!th]
\caption{Performance comparison on the RSA module. The best results are in
bold. \label{tab:RSA}}

\centering{}\centering{}\centering{}\doublerulesep=0.5pt \resizebox{0.48\textwidth}{!}{%
\begin{tabular}{ccccc}
\hline
\multirow{2}{*}{} & \multirow{2}{*}{Methods} & \multicolumn{3}{c}{F-Score}\tabularnewline
\cline{3-5} \cline{4-5} \cline{5-5}
 &  & GSV  &  & TSUNAMI\tabularnewline
\hline
(g)  & Dense Fusion (PCF)  & 0.755  &  & 0.857\tabularnewline
(h)  & Dense Fusion (PCF) + RSA  & \textbf{0.761}  &  & \textbf{0.861}\tabularnewline
\hline
\end{tabular}}
\end{table}

\textbf{Effects of RSA modules.} To evaluate the benefits of the proposed
RSA modules, we also re-implement our approach with/without them.
Tab. \ref{tab:RSA} shows the quantitative performance. In terms of
F-Score, our network with RSA modules achieves 0.761/0.861 on GSV/TSUNAMI
datasets. Fig. \ref{fig:A-comparison-on} shows the attention masks
for a typical street scene pair. We convert the attention masks to
heat maps for better visualization. One can see that the highlighted
regions by our RSA module are coarsely consistent with the changed
regions.

\begin{table}[!th]
\centering{}\centering{}\caption{The parameters of MPFL and replaced convolutions. $(k,m,n,c)$ represent
the kernel size, input channel number, out channel number and group
number, respectively.\label{tab:The-parameter-details}}
\centering{}\doublerulesep=0.5pt \resizebox{0.48\textwidth}{!}{%
\begin{tabular}{cccc}
\hline
Location  & No. of Parameters  & MPFL  & Substituted Convolutional Layer\tabularnewline
\hline
Conv2\_2  & 0.02 MB  & (128, 64)  & (3, 128, 64, 4)\tabularnewline
Conv3\_3  & 0.08 MB  & (256, 128)  & (3, 256, 128, 4)\tabularnewline
Conv4\_3  & 0.34 MB  & (512, 256)  & (3, 512, 256, 4)\tabularnewline
Conv5\_3  & 0.34 MB  & (512, 256)  & (3, 512, 256, 4)\tabularnewline
\hline
\end{tabular}}
\end{table}

\begin{table}[!th]
\caption{Performance comparison on the MPFL strategy. The best results are
in bold. \label{tab:Performance-comparison-of-2}}

\centering{}\centering{}\centering{}\doublerulesep=0.5pt \resizebox{0.48\textwidth}{!}{%
\begin{tabular}{ccccc}
\hline
\multirow{2}{*}{} & \multirow{2}{*}{Methods} & \multicolumn{3}{c}{F-Score}\tabularnewline
\cline{3-5} \cline{4-5} \cline{5-5}
 &  & GSV  &  & TSUNAMI\tabularnewline
\hline
(i)  & Dense Fusion (PCF) + RSA + Conv  & 0.764  &  & 0.862\tabularnewline
(j)  & Dense Fusion (PCF) + RSA + 2Conv  & 0.762  &  & 0.859\tabularnewline
(k)  & Dense Fusion (PCF) + RSA + MPFL $-$ partition  & 0.768  &  & 0.860\tabularnewline
($l$)  & Dense Fusion (PCF) + RSA + MPFL  & \textbf{0.776}  &  & \textbf{0.868}\tabularnewline
\hline
\end{tabular}}
\end{table}

\textbf{Effects of the MPFL strategy.} The proposed MPFL strategy
aims to extract local and global features for adapting to the diverse
locations of changed regions.\textcolor{red}{{} }One could suspect that
the performance improvement might actually be due to the introduction
of additional parameters. For example, whether the performance could
also simply be achieved by a convolutional layer with a comparable
number of parameters.
\begin{figure*}[!tp]
\begin{centering}
\includegraphics[width=0.96\linewidth,height=20cm]{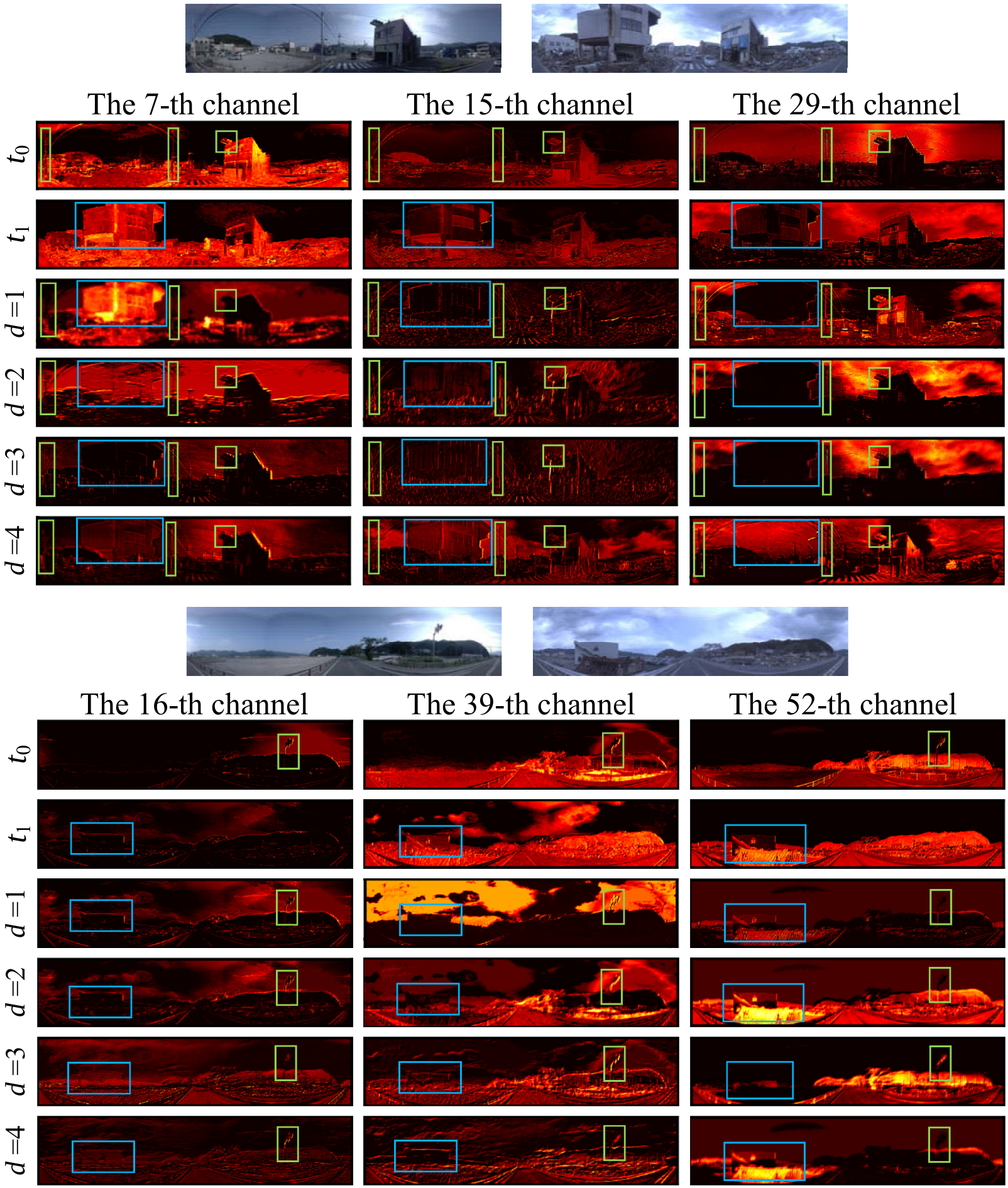}
\par\end{centering}
\caption{Illustration of PAGC feature maps. For better visualization, we sample
several channels from the PAGC module at the Conv1\_2 layer. The first
row of each sub-figure shows the raw images at $t_{0}$ and $t_{1}$.
The second and third rows of each sub-figure show cross feature maps
from the paired images. The 4th-7th rows are feature maps generated
with different dilation rates $d$. The changed regions at $t_{0}$
and $t_{1}$ are marked by green and blue bounding boxes, respectively.
\label{fig:Sampled-feature-maps}}
\vspace{-4mm}

\end{figure*}

To address above concerns, we replace the MPFL with convolutional
layers which have comparable parameters. More specifically, our MPFL
modules at different levels have different parameters. Thus, we use
specific convolutional layers to replace the MPFL at each level. Tab.
\ref{tab:The-parameter-details} shows structure details. While Tab.
\ref{tab:Performance-comparison-of-2} reports the results with different
settings. We observe the following facts: 1) With only one convolutional
layer with comparable parameters, the resulting model (model i) achieves
0.764/0.862 F-Score on the GSV/TSUNAMI datasets, respectively. These
results are worse than our MPFL strategy (model $l$) which achieves
0.776/0.868. This indicates that our MPFL strategy is more effective.
2) With two additional convolutional layers (model j), the network
achieves 0.762/0.859, which are worse than the single convolutional
layer (model i). The fact indicates that simply adding parameters
is not an effective way for the SSCD task.

\textbf{Effects of the Partition in MPFL.} To demonstrate the effect
of Partition, we conducted an ablation experiment by removing the
partition step but keeping different-sized convolutions. Note that
without partition, each branch contains the intact feature map, hence
we remove the multiple independent convolutions and replace it by
a single convolution with comparable parameters. For example, four
$1\times1$ convolutions in MPFL (Branch 1) are changed into one $2\times2$
convolution in the added ablation experiment. As shown in Tab. \ref{tab:Performance-comparison-of-2},
model without Partition (model k) achieves F-Score of 0.768/0.860
on GSV/TSUNAMI, which is worse than MPFL. MPFL which utilizes partition
process has a more powerful capacity of feature learning.

\textbf{Effects of PAGC.} As a key module in PCF, the PAGC plays an
important role in capturing the multi-scale correlation between channels.
As mentioned in Section III, each channel pair in the group of PAGC
contains the information of changed regions. As shown in Fig. \ref{fig:Sampled-feature-maps},
the fused feature maps (from the 4-th row to the 7-th in each sub-figure)
contain the information of input paired features. Besides, using different
dilation rates is able to extract multi-scale feature representations,
which enriches the fusion information. These examples confirm the
reasonability of our strategy: (1) Fusing features in a channel-wise
manner incorporates the information of each paired channels, and enhances
the expression of changes at each fused channel. (2) Using different
dilation rates enriches the representations of the multi-scale information.

\section{Conclusion\label{sec:Conclusion}}

In this paper, we propose a novel deep learning framework, named HPCFNet,
for the SSCD task. To enhance the feature fusion, we introduce a PCF
module to hierarchically fuse feature maps in a channel-wise manner.
Next, we propose a RSA module to adaptively highlight the features
indicating the changed regions. To enrich the scene information, we
also propose a MPFL strategy to extract features in a global-to-local
manner. Extensive experiments on three publicly available SSCD datasets
(i.e., PCD, VL-CMU-CD and CDnet2014) show that our approach achieves
remarkable performance compared to other state-of-the-art methods.
\ifCLASSOPTIONcaptionsoff
  \newpage
\fi
\bibliographystyle{IEEEtran}
\bibliography{IEEEabrv,refs}

\begin{thebibliography}{10}
\providecommand{\url}[1]{#1}
\csname url@samestyle\endcsname
\providecommand{\newblock}{\relax}
\providecommand{\bibinfo}[2]{#2}
\providecommand{\BIBentrySTDinterwordspacing}{\spaceskip=0pt\relax}
\providecommand{\BIBentryALTinterwordstretchfactor}{4}
\providecommand{\BIBentryALTinterwordspacing}{\spaceskip=\fontdimen2\font plus
\BIBentryALTinterwordstretchfactor\fontdimen3\font minus
  \fontdimen4\font\relax}
\providecommand{\BIBforeignlanguage}[2]{{%
\expandafter\ifx\csname l@#1\endcsname\relax
\typeout{** WARNING: IEEEtran.bst: No hyphenation pattern has been}%
\typeout{** loaded for the language `#1'. Using the pattern for}%
\typeout{** the default language instead.}%
\else
\language=\csname l@#1\endcsname
\fi
#2}}
\providecommand{\BIBdecl}{\relax}
\BIBdecl

\bibitem{wu2017kernel}
C.~Wu, L.~Zhang, and B.~Du, ``Kernel slow feature analysis for scene change
  detection,'' \emph{IEEE Transactions on Geoscience and Remote Sensing
  (TGRS)}, vol.~55, no.~4, pp. 2367--2384, 2017.

\bibitem{wu2016scene}
C.~Wu, L.~Zhang, and L.~Zhang, ``A scene change detection framework for
  multi-temporal very high resolution remote sensing images,'' \emph{Signal
  Processing}, vol. 124, pp. 184--197, 2016.

\bibitem{hussain2013change}
M.~Hussain, D.~Chen, A.~Cheng, H.~Wei, and D.~Stanley, ``Change detection from
  remotely sensed images: From pixel-based to object-based approaches,''
  \emph{ISPRS Journal of Photogrammetry and Remote Sensing}, vol.~80, pp.
  91--106, 2013.

\bibitem{mou2018learning}
L.~Mou, L.~Bruzzone, and X.~X. Zhu, ``Learning spectral-spatial-temporal
  features via a recurrent convolutional neural network for change detection in
  multispectral imagery,'' \emph{IEEE Transactions on Geoscience and Remote
  Sensing (TGRS)}, vol.~57, no.~2, pp. 924--935, 2018.

\bibitem{du2018unsupervised}
B.~Du, Y.~Wang, C.~Wu, and L.~Zhang, ``Unsupervised scene change detection via
  latent dirichlet allocation and multivariate alteration detection,''
  \emph{IEEE Journal of Selected Topics in Applied Earth Observations and
  Remote Sensing (JSTARS)}, vol.~11, no.~12, pp. 4676--4689, 2018.

\bibitem{wu2013slow}
C.~Wu, B.~Du, and L.~Zhang, ``Slow feature analysis for change detection in
  multispectral imagery,'' \emph{IEEE Transactions on Geoscience and Remote
  Sensing (TGRS)}, vol.~52, no.~5, pp. 2858--2874, 2013.

\bibitem{sakurada2017dense}
K.~Sakurada, W.~Wang, N.~Kawaguchi, and R.~Nakamura, ``Dense optical flow based
  change detection network robust to difference of camera viewpoints,''
  \emph{arXiv:1712.02941}, 2017.

\bibitem{sakurada2018weakly}
K.~Sakurada, ``Weakly supervised silhouette-based semantic change detection,''
  \emph{arXiv:1811.11985}, 2018.

\bibitem{guo2018learning}
E.~Guo, X.~Fu, J.~Zhu, M.~Deng, Y.~Liu, Q.~Zhu, and H.~Li, ``Learning to
  measure change: Fully convolutional siamese metric networks for scene change
  detection,'' \emph{arXiv:1810.09111}, 2018.

\bibitem{khan2016learning}
S.~H. Khan, X.~He, F.~Porikli, M.~Bennamoun, F.~Sohel, and R.~Togneri,
  ``Learning deep structured network for weakly supervised change detection,''
  \emph{arXiv:1606.02009}, 2016.

\bibitem{sakurada2015change}
K.~Sakurada and T.~Okatani, ``Change detection from a street image pair using
  cnn features and superpixel segmentation,'' in \emph{Proceedings of the
  British Machine Vision Conference (BMVC)}, 2015, pp. 61--73.

\bibitem{rosin1998thresholding}
P.~Rosin, ``Thresholding for change detection,'' in \emph{Proceedings of the
  IEEE International Conference on Computer Vision (ICCV)}, 1998, pp. 274--279.

\bibitem{rosin2003evaluation}
P.~L. Rosin and E.~Ioannidis, ``Evaluation of global image thresholding for
  change detection,'' \emph{Pattern Recognition Letters (PRL)}, vol.~24,
  no.~14, pp. 2345--2356, 2003.

\bibitem{pollard2007change}
T.~Pollard and J.~L. Mundy, ``Change detection in a 3-d world,'' in
  \emph{Proceedings of the IEEE Computer Vision and Pattern Recognition
  (CVPR)}, 2007, pp. 1--6.

\bibitem{schindler2010probabilistic}
G.~Schindler and F.~Dellaert, ``Probabilistic temporal inference on
  reconstructed 3d scenes,'' in \emph{Proceedings of the IEEE Computer Vision
  and Pattern Recognition (CVPR)}, 2010, pp. 1410--1417.

\bibitem{taneja2011image}
A.~Taneja, L.~Ballan, and M.~Pollefeys, ``Image based detection of geometric
  changes in urban environments,'' in \emph{Proceedings of IEEE International
  Conference on Computer Vision (ICCV)}, 2011, pp. 2336--2343.

\bibitem{taneja2013city}
------, ``City-scale change detection in cadastral 3{D} models using images,''
  in \emph{Proceedings of the IEEE Computer Vision and Pattern Recognition
  (CVPR)}, 2013, pp. 113--120.

\bibitem{long2015fully}
J.~Long, E.~Shelhamer, and T.~Darrell, ``Fully convolutional networks for
  semantic segmentation,'' in \emph{Proceedings of the IEEE Computer Vision and
  Pattern Recognition (CVPR)}, 2015, pp. 3431--3440.

\bibitem{ronneberger2015u}
O.~Ronneberger, P.~Fischer, and T.~Brox, ``U-net: Convolutional networks for
  biomedical image segmentation,'' in \emph{Proceedings of the International
  Conference on Medical Image Computing and Computer-Assisted Intervention
  (MICCAI)}.\hskip 1em plus 0.5em minus 0.4em\relax Springer, 2015, pp.
  234--241.

\bibitem{szegedy2015going}
C.~Szegedy, W.~Liu, Y.~Jia, P.~Sermanet, S.~Reed, D.~Anguelov, D.~Erhan,
  V.~Vanhoucke, and A.~Rabinovich, ``Going deeper with convolutions,'' in
  \emph{Proceedings of the IEEE Computer Vision and Pattern Recognition
  (CVPR)}, 2015, pp. 1--9.

\bibitem{alcantarilla2018street}
P.~F. Alcantarilla, S.~Stent, G.~Ros, R.~Arroyo, and R.~Gherardi, ``Street-view
  change detection with deconvolutional networks,'' \emph{Autonomous Robots},
  vol.~42, no.~7, pp. 1301--1322, 2018.

\bibitem{zhang2020rapnet}
P.~Zhang, W.~Liu, Y.~Lei, H.~Wang, and H.~Lu, ``Rapnet: Residual atrous pyramid
  network for importance-aware street scene parsing,'' \emph{IEEE Transactions
  on Image Processing (TIP)}, vol.~29, pp. 5010--5021, 2020.

\bibitem{wang2014cdnet}
Y.~Wang, P.-M. Jodoin, F.~Porikli, J.~Konrad, Y.~Benezeth, and P.~Ishwar,
  ``Cdnet 2014: an expanded change detection benchmark dataset,'' in
  \emph{Proceedings of the IEEE Computer Vision and Pattern Recognition
  Workshops (CVPRW)}, 2014, pp. 387--394.

\bibitem{zhang2020deep}
P.~Zhang, W.~Liu, Y.~Lei, H.~Wang, and H.~Lu, ``Deep multiphase level set for
  scene parsing,'' \emph{IEEE Transactions on Image Processing (TIP)}, vol.~29,
  pp. 4556--4567, 2020.

\bibitem{zhang2019hyperfusion}
P.~Zhang, W.~Liu, Y.~Lei, and H.~Lu, ``Hyperfusion-net: Hyper-densely
  reflective feature fusion for salient object detection,'' \emph{Pattern
  Recognition (PR)}, vol.~93, pp. 521--533, 2019.

\bibitem{zhang2019cascaded}
P.~Zhang, W.~Liu, Y.~Lei, H.~Lu, and X.~Yang, ``Cascaded context pyramid for
  full-resolution 3d semantic scene completion,'' in \emph{Proceedings of the
  IEEE International Conference on Computer Vision (ICCV)}, 2019, pp.
  7801--7810.

\bibitem{zhang2018salient}
P.~Zhang, W.~Liu, H.~Lu, and C.~Shen, ``Salient object detection by lossless
  feature reflection,'' \emph{International Joint Conference on Artificial
  Intelligence (IJCAI)}, vol.~14, no.~4, pp. 1149--1155, 2018.

\bibitem{ma2020global}
Y.~Ma, Y.~Guo, H.~Liu, Y.~Lei, and G.~Wen, ``Global context reasoning for
  semantic segmentation of 3d point clouds,'' in \emph{Proceedings of the IEEE
  Winter Conference on Applications of Computer Vision (WACV)}, 2020, pp.
  2931--2940.

\bibitem{radke2005image}
R.~J. Radke, S.~Andra, O.~Al-Kofahi, and B.~Roysam, ``Image change detection
  algorithms: a systematic survey,'' \emph{IEEE Transactions on Image
  Processing (TIP)}, vol.~14, no.~3, pp. 294--307, 2005.

\bibitem{he2016deep}
K.~He, X.~Zhang, S.~Ren, and J.~Sun, ``Deep residual learning for image
  recognition,'' in \emph{Proceedings of the IEEE Computer Vision and Pattern
  Recognition (CVPR)}, 2016, pp. 770--778.

\bibitem{krizhevsky2012imagenet}
A.~Krizhevsky, I.~Sutskever, and G.~E. Hinton, ``Imagenet classification with
  deep convolutional neural networks,'' in \emph{Proceedings of the Advances in
  Neural Information Processing Systems (NuerIPS)}, 2012, pp. 1097--1105.

\bibitem{lecun1998gradient}
Y.~LeCun, L.~Bottou, Y.~Bengio, P.~Haffner \emph{et~al.}, ``Gradient-based
  learning applied to document recognition,'' \emph{Proceedings of the IEEE},
  vol.~86, no.~11, pp. 2278--2324, 1998.

\bibitem{simonyan2014very}
K.~Simonyan and A.~Zisserman, ``Very deep convolutional networks for
  large-scale image recognition,'' \emph{arXiv:1409.1556}, 2014.

\bibitem{liu2020semi}
Y.~Liu, L.~Liu, P.~Wang, P.~Zhang, and Y.~Lei, ``Semi-supervised crowd counting
  via self-training on surrogate tasks,'' in \emph{Proceedings of the European
  Conference on Computer Vision (ECCV)}.\hskip 1em plus 0.5em minus 0.4em\relax
  Springer, 2020.

\bibitem{lei2021towards}
Y.~Lei, Y.~Liu, P.~Zhang, and L.~Liu, ``Towards using count-level weak
  supervision for crowd counting,'' \emph{Pattern Recognition (PR)}, vol. 109,
  p. 107616, 2021.

\bibitem{zhan2017change}
Y.~Zhan, K.~Fu, M.~Yan, X.~Sun, H.~Wang, and X.~Qiu, ``Change detection based
  on deep siamese convolutional network for optical aerial images,'' \emph{IEEE
  Geoscience and Remote Sensing Letters (GRSL)}, vol.~14, no.~10, pp.
  1845--1849, 2017.

\bibitem{zhang2019salient}
P.~Zhang, W.~Liu, H.~Lu, and C.~Shen, ``Salient object detection with lossless
  feature reflection and weighted structural loss,'' \emph{IEEE Transactions on
  Image Processing (TIP)}, vol.~28, no.~6, pp. 3048--3060, 2019.

\bibitem{itti2001computational}
L.~Itti and C.~Koch, ``Computational modelling of visual attention,''
  \emph{Nature Reviews Neuroscience}, vol.~2, no.~3, p. 194, 2001.

\bibitem{itti1998model}
L.~Itti, C.~Koch, and E.~Niebur, ``A model of saliency-based visual attention
  for rapid scene analysis,'' \emph{IEEE Transactions on Pattern Analysis and
  Machine Intelligence (TPAMI)}, no.~11, pp. 1254--1259, 1998.

\bibitem{olshausen1993neurobiological}
B.~A. Olshausen, C.~H. Anderson, and D.~C. Van~Essen, ``A neurobiological model
  of visual attention and invariant pattern recognition based on dynamic
  routing of information,'' \emph{Journal of Neuroscience}, vol.~13, no.~11,
  pp. 4700--4719, 1993.

\bibitem{zhang2019deep}
P.~Zhang, W.~Liu, H.~Wang, Y.~Lei, and H.~Lu, ``Deep gated attention networks
  for large-scale street-level scene segmentation,'' \emph{Pattern Recognition
  (PR)}, vol.~88, pp. 702--714, 2019.

\bibitem{larochelle2010learning}
H.~Larochelle and G.~E. Hinton, ``Learning to combine foveal glimpses with a
  third-order boltzmann machine,'' in \emph{Proceedings of the Advances in
  Neural Information Processing Systems (NuerIPS)}, 2010, pp. 1243--1251.

\bibitem{mnih2014recurrent}
V.~Mnih, N.~Heess, A.~Graves \emph{et~al.}, ``Recurrent models of visual
  attention,'' in \emph{Proceedings of the Advances in Neural Information
  Processing Systems (NuerIPS)}, 2014, pp. 2204--2212.

\bibitem{jaderberg2015spatial}
M.~Jaderberg, K.~Simonyan, A.~Zisserman \emph{et~al.}, ``Spatial transformer
  networks,'' in \emph{Proceedings of the Advances in Neural Information
  Processing Systems (NuerIPS)}, 2015, pp. 2017--2025.

\bibitem{fu2019dual}
J.~Fu, J.~Liu, H.~Tian, Y.~Li, Y.~Bao, Z.~Fang, and H.~Lu, ``Dual attention
  network for scene segmentation,'' in \emph{Proceedings of the IEEE Conference
  on Computer Vision and Pattern Recognition (CVPR)}, 2019, pp. 3146--3154.

\bibitem{chen2018reverse}
S.~Chen, X.~Tan, B.~Wang, and X.~Hu, ``Reverse attention for salient object
  detection,'' in \emph{Proceedings of the European Conference on Computer
  Vision (ECCV)}.\hskip 1em plus 0.5em minus 0.4em\relax Springer, 2018, pp.
  234--250.

\bibitem{cao2015look}
C.~Cao, X.~Liu, Y.~Yang, Y.~Yu, J.~Wang, Z.~Wang, Y.~Huang, L.~Wang, C.~Huang,
  W.~Xu \emph{et~al.}, ``Look and think twice: Capturing top-down visual
  attention with feedback convolutional neural networks,'' in \emph{Proceedings
  of the IEEE Computer Vision and Pattern Recognition (CVPR)}, 2015, pp.
  2956--2964.

\bibitem{hu2018squeeze}
J.~Hu, L.~Shen, and G.~Sun, ``Squeeze-and-excitation networks,'' in
  \emph{Proceedings of the IEEE Computer Vision and Pattern Recognition
  (CVPR)}, 2018, pp. 7132--7141.

\bibitem{woo2018cbam}
S.~Woo, J.~Park, J.-Y. Lee, and I.~So~Kweon, ``Cbam: Convolutional block
  attention module,'' in \emph{Proceedings of the European Conference on
  Computer Vision (ECCV)}.\hskip 1em plus 0.5em minus 0.4em\relax Springer,
  2018, pp. 3--19.

\bibitem{chen2017deeplab}
L.-C. Chen, G.~Papandreou, I.~Kokkinos, K.~Murphy, and A.~L. Yuille, ``Deeplab:
  Semantic image segmentation with deep convolutional nets, atrous convolution,
  and fully connected crfs,'' \emph{IEEE Transactions on Pattern Analysis and
  Machine Intelligence (TPAMI)}, vol.~40, no.~4, pp. 834--848, 2017.

\bibitem{ioffe2015batch}
S.~Ioffe and C.~Szegedy, ``Batch normalization: Accelerating deep network
  training by reducing internal covariate shift,'' in \emph{International
  Conference on Machine Learning (ICML)}, 2015, pp. 448--456.

\bibitem{deng2009imagenet}
J.~Deng, W.~Dong, R.~Socher, L.-J. Li, K.~Li, and L.~Fei-Fei, ``Imagenet: A
  large-scale hierarchical image database,'' in \emph{Proceedings of the IEEE
  Computer Vision and Pattern Recognition (CVPR)}, 2009, pp. 248--255.

\bibitem{paszke2017automatic}
A.~Paszke, S.~Gross, S.~Chintala, G.~Chanan, E.~Yang, Z.~DeVito, Z.~Lin,
  A.~Desmaison, L.~Antiga, and A.~Lerer, ``Automatic differentiation in
  pytorch,'' 2017.

\bibitem{tola2009daisy}
E.~Tola, V.~Lepetit, and P.~Fua, ``Daisy: An efficient dense descriptor applied
  to wide-baseline stereo,'' \emph{IEEE Transactions on Pattern Analysis and
  Machine Intelligence (TPAMI)}, vol.~32, no.~5, pp. 815--830, 2009.

\bibitem{kim2015dasc}
S.~Kim, D.~Min, B.~Ham, S.~Ryu, M.~N. Do, and K.~Sohn, ``Dasc: Dense adaptive
  self-correlation descriptor for multi-modal and multi-spectral
  correspondence,'' in \emph{Proceedings of the IEEE Computer Vision and
  Pattern Recognition (CVPR)}, 2015, pp. 2103--2112.

\bibitem{glorot2010understanding}
X.~Glorot and Y.~Bengio, ``Understanding the difficulty of training deep
  feedforward neural networks,'' in \emph{Proceedings of the International
  Conference on Artificial Intelligence and Statistics (AISTATS)}.\hskip 1em
  plus 0.5em minus 0.4em\relax Citeseer, 2010, pp. 249--256.

\bibitem{lowe2004distinctive}
D.~G. Lowe, ``Distinctive image features from scale-invariant keypoints,''
  \emph{International Journal of Computer Vision (IJCV)}, vol.~60, no.~2, pp.
  91--110, 2004.

\bibitem{st2014subsense}
P.-L. St-Charles, G.-A. Bilodeau, and R.~Bergevin, ``Subsense: A universal
  change detection method with local adaptive sensitivity,'' \emph{IEEE
  Transactions on Image Processing (TIP)}, vol.~24, no.~1, pp. 359--373, 2014.

\bibitem{babaee2018deep}
M.~Babaee, D.~T. Dinh, and G.~Rigoll, ``A deep convolutional neural network for
  video sequence background subtraction,'' \emph{Pattern Recognition (PR)},
  vol.~76, pp. 635--649, 2018.

\bibitem{braham2017semantic}
M.~Braham, S.~Pi{\'e}rard, and M.~Van~Droogenbroeck, ``Semantic background
  subtraction,'' in \emph{Proceedings of the IEEE International Conference on
  Image Processing (ICIP)}, 2017, pp. 4552--4556.

\bibitem{bianco2017combination}
S.~Bianco, G.~Ciocca, and R.~Schettini, ``Combination of video change detection
  algorithms by genetic programming,'' \emph{IEEE Transactions on Evolutionary
  Computation (TEC)}, vol.~21, no.~6, pp. 914--928, 2017.

\bibitem{liang2015co}
D.~Liang, M.~Hashimoto, K.~Iwata, X.~Zhao \emph{et~al.}, ``Co-occurrence
  probability-based pixel pairs background model for robust object detection in
  dynamic scenes,'' \emph{Pattern Recognition (PR)}, vol.~48, no.~4, pp.
  1374--1390, 2015.

\bibitem{wang2017interactive}
Y.~Wang, Z.~Luo, and P.-M. Jodoin, ``Interactive deep learning method for
  segmenting moving objects,'' \emph{Pattern Recognition Letters (PRL)},
  vol.~96, pp. 66--75, 2017.

\bibitem{lim2020learning}
L.~A. Lim and H.~Y. Keles, ``Learning multi-scale features for foreground
  segmentation,'' \emph{Pattern Analysis and Applications}, vol.~23, no.~3, pp.
  1369--1380, 2020.

\bibitem{zheng2020novel}
W.~Zheng, K.~Wang, and F.-Y. Wang, ``A novel background subtraction algorithm
  based on parallel vision and bayesian gans,'' \emph{Neurocomputing}, vol.
  394, pp. 178--200, 2020.

\end{thebibliography}
\end{document}